\documentclass[runningheads]{llncs}

\usepackage{eccv}

\usepackage{eccvabbrv}
\usepackage{graphicx}
\usepackage{booktabs}
\usepackage[accsupp]{axessibility}  % Improves PDF readability for those with disabilities.
\usepackage[pagebackref,breaklinks,colorlinks]{hyperref}
\usepackage{orcidlink}
\usepackage{url}   
\usepackage{listings}% simple URL typesetting
\lstdefinestyle{mypython}{language=Python, basicstyle=\ttfamily\small, keywordstyle=\color{blue}, commentstyle=\color{gray}, stringstyle=\color{red}, showstringspaces=false, breaklines=true, frame=none}
\usepackage{booktabs}       % professional-quality tables
\usepackage{amsfonts}       % blackboard math symbols
\usepackage{nicefrac}       % compact symbols for 1/2, etc.
\usepackage{microtype}      % microtypography
\usepackage{xcolor}         % colors
\usepackage{makecell}
\usepackage{epsfig}
\usepackage{graphicx}
\usepackage{amsmath}
\usepackage{amssymb}
\usepackage{float}

\usepackage{amsmath}
\usepackage{amssymb}

\definecolor{accent}{RGB}{31,78,121}
\definecolor{codebg}{RGB}{245,245,245}
\definecolor{codecomment}{RGB}{100,100,100}
\definecolor{codestring}{RGB}{140,60,30}
\definecolor{codekw}{RGB}{31,78,121}
\definecolor{darkgray}{RGB}{80,80,80}
\definecolor{green_box}{RGB}{0,128,0}
\definecolor{red_box}{RGB}{180,0,0}
\definecolor{blue_box}{RGB}{0,0,180}
 
\newcommand{\Pc}{P_c}
\newcommand{\Tmax}{T_{\max}}

\newcommand{\greenbox}{\textcolor{green_box}{\textbf{green}}}
\newcommand{\redbox}{\textcolor{red_box}{\textbf{red}}}
\newcommand{\bluebox}{\textcolor{blue_box}{\textbf{blue}}}

\usepackage{colortbl}
\usepackage{enumitem}
\usepackage{multirow}
\usepackage{comment}
\usepackage{textcomp,booktabs}
\usepackage[font=tiny]{subcaption}
\usepackage[mathscr]{eucal}
\usepackage[export]{adjustbox}
\usepackage{multicol,lipsum}
\usepackage{arydshln}
\usepackage{pifont}
\usepackage{booktabs}
\usepackage{bbding}

\usepackage{algpseudocode}
\def\0{{\bf 0}}
\def\1{{\bf 1}}

\newcommand{\cmark}{\textcolor{darkgreen}{\ding{51}}}%
\newcommand{\xmark}{\textcolor{red}{\ding{55}}}%

\definecolor{purple}{rgb}{0.56,0.27,0.68}
\definecolor{red}{rgb}{0.95,0.4,0.4}
\definecolor{purered}{rgb}{1,0,0}
\definecolor{blue}{rgb}{0.4,0.4,0.95}
\definecolor{darkblue}{rgb}{0,0,0.8}
\definecolor{grey}{rgb}{0.6,0.6,0.6}
\definecolor{col1}{RGB}{232, 161, 148}
\definecolor{col11}{RGB}{255, 228, 228}
\definecolor{col2}{RGB}{148, 187, 232}
\definecolor{col33}{RGB}{206, 239, 255}
\definecolor{col3}{RGB}{233, 255, 245}
\definecolor{lightgrey}{rgb}{0.85,0.85,0.85}
\definecolor{lightlightgrey}{rgb}{0.9,0.9,0.9}
\definecolor{verylightBG}{rgb}{0.9,0.99,0.99}
\definecolor{darkgreen}{rgb}{0., 0.85, 0.5}
\definecolor{lightgray}{gray}{0.75}

\definecolor{gtred}{RGB}{204, 0, 0}
\definecolor{predgreen}{RGB}{31, 237, 31}
\definecolor{figGreen}{RGB}{56, 118, 29}
\definecolor{lightgray}{gray}{0.75}

\graphicspath{{./figures/}}   %  image

\usepackage{algorithm2e}
\usepackage{pythonhighlight} 
\usepackage{fontawesome5}
\lstnewenvironment{pythonic}[1][]{\lstset{style=mypython, frame=none, #1}}{}
\usepackage[most]{tcolorbox}

\newcommand{\greybox}[1]{%
  \vspace{0.8\baselineskip}
  \begin{tcolorbox}[
    colback=gray!10,
    colframe=gray!40,
    arc=6pt,
    boxrule=0pt,
    left=8pt,
    right=8pt,
    top=6pt,
    bottom=6pt,
  ]
  #1
  \end{tcolorbox}
  \vspace{0.8\baselineskip}
}

\usepackage[capitalize]{cleveref}
\crefname{section}{Sec.}{Secs.}
\Crefname{section}{Section}{Sections}
\Crefname{table}{Table}{Tables}
\crefname{table}{Tab.}{Tabs.}

\graphicspath{{./figures/}}

\begin{document}

\title{DetPO: In-Context Learning with Multi-Modal LLMs for Few-Shot Object Detection} 

\titlerunning{In-Context Learning with Multi-Modal LLMs for Few-Shot Object Detection}

\author{Gautam Rajendrakumar Gare$^{1,}$\thanks{Equal Contribution} \and Neehar Peri$^{1,\star}$ \and Matvei Popov$^2$ \and Shruti Jain$^1$ \and John Galeotti$^1$ \and Deva Ramanan$^1$}

\authorrunning{Gare and Peri et. al.}

\institute{Carnegie Mellon University \and Roboflow}

\maketitle

\begin{abstract}
Multi-Modal LLMs (MLLMs) demonstrate strong visual grounding capabilities on popular object detection benchmarks like OdinW-13 and RefCOCO. However, state-of-the-art models still struggle to generalize to out-of-distribution classes, tasks and imaging modalities not typically found in their pre-training. While in-context prompting is a common strategy to improve performance across diverse tasks, we find that it often yields lower detection accuracy than prompting with class names alone. This suggests that current MLLMs cannot yet effectively leverage few-shot visual examples and rich textual descriptions for object detection. Since frontier MLLMs are typically only accessible via APIs, and state-of-the-art open-weights models are prohibitively expensive to fine-tune on consumer-grade hardware, we instead explore black-box prompt optimization for few-shot object detection. To this end, we propose Detection Prompt Optimization (DetPO), a gradient-free test-time optimization approach that refines text-only prompts by maximizing detection accuracy on few-shot visual training examples while calibrating prediction confidence. Our proposed approach yields consistent improvements across generalist MLLMs on Roboflow20-VL and LVIS, outperforming prior black-box approaches by up to 9.7 mAP. Our code and optimized prompts are available on our \href{https://ggare-cmu.github.io/DetPO/}{project page}.

\end{abstract}

\begin{figure}[t]
    \centering
    \includegraphics[width=\linewidth]{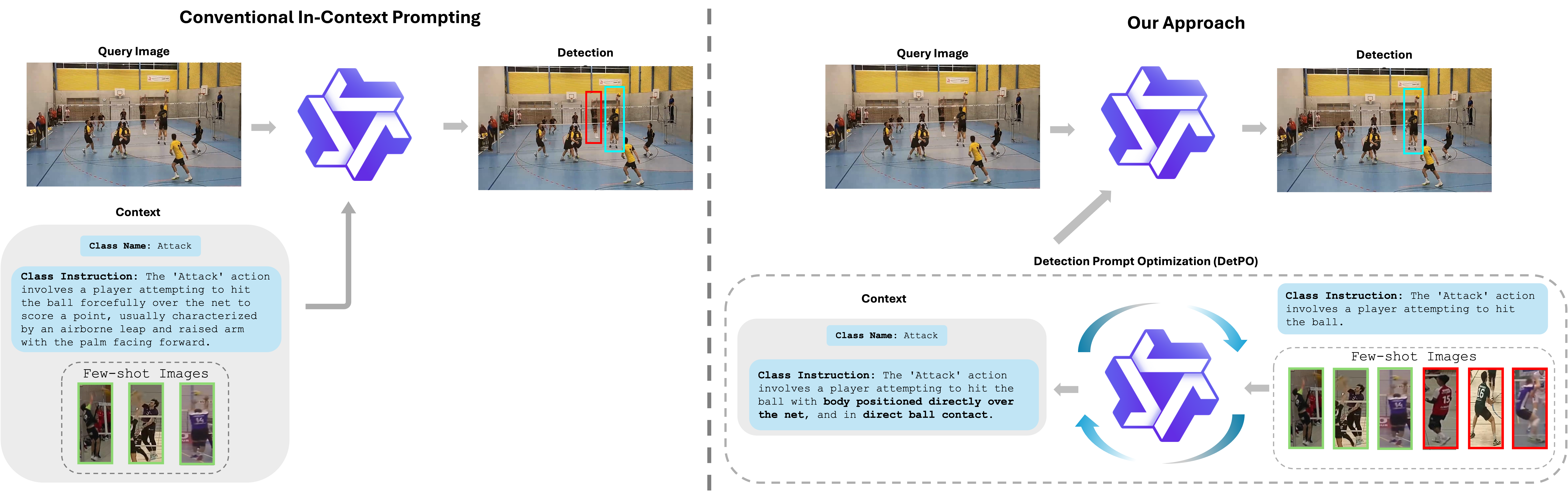}
    \caption{{\bf In-Context Learning for Object Detection.} We cast the problem of gradient-free few-shot object detection as multimodal in-context learning (ICL). Here, a frozen multi-modal LLM (MLLM) is presented with a class name, a textual description, and a few visual examples ({\bf left}), similar to the instructions given to a human annotator tasked with annotating that class~\cite{robicheaux2025roboflow100}. Rather than presenting the visual examples directly to the MLLM, we find that it is far more effective to use them to optimize a better text-only prompt ({\bf right}) via prompt optimization; we use a critique MLLM to discover prompt instructions that perform better on the few-shot training dataset. These improved instructions are then fed into the target MLLM. In practice, we use the same MLLM for both the critique and target.
    }
    \label{fig:teaser}
\end{figure}

{
\setlength{\tabcolsep}{1.0em}
\begin{table}[t]
\centering
\caption{\textbf{Multi-Modal ICL Hurts Detection Accuracy}. Current MLLMs struggle to learn from multi-modal in-context examples for object detection. We posit that rigid post-training prompt structures make it difficult to effectively leverage additional context. To address this limitation, we propose Detection Prompt Optimization (DetPO) to encode few-shot multi-modal examples into a single text-only prompt, significantly improving detection accuracy.
} 
\label{tab:naive_icl}
\scalebox{0.90}{
\begin{tabular}{|lcccc|}
\hline
\rowcolor{gray!10}
\multicolumn{1}{|l|}{\textbf{Method}}                     & \multicolumn{1}{c|}{\textbf{Class Names}} & \multicolumn{1}{c|}{\textbf{Instructions}} & \multicolumn{1}{c|}{\textbf{Images}} & \textbf{mAP} \\ \hline
\multicolumn{1}{|l|}{Qwen2.5-VL \cite{bai2025qwen2} (7B)}      & \multicolumn{1}{c|}{\cmark}      & \multicolumn{1}{c|}{\xmark}      & \multicolumn{1}{c|}{\xmark}  &    4.6         \\ \hline
\multicolumn{1}{|l|}{Qwen2.5-VL \cite{bai2025qwen2} (7B)}      & \multicolumn{1}{c|}{\cmark}      & \multicolumn{1}{c|}{\cmark}      & \multicolumn{1}{c|}{\xmark}  &       \textbf{6.2}      \\ \hline
\multicolumn{1}{|l|}{Qwen2.5-VL \cite{bai2025qwen2} (7B)}      & \multicolumn{1}{c|}{\cmark}      & \multicolumn{1}{c|}{\cmark}      & \multicolumn{1}{c|}{\cmark}  &       1.8      \\ \hline \hline
\multicolumn{1}{|l|}{Qwen2.5-VL \cite{bai2025qwen2} (72B)}      & \multicolumn{1}{c|}{\cmark}      & \multicolumn{1}{c|}{\xmark}      & \multicolumn{1}{c|}{\xmark}  &     7.1        \\ \hline
\multicolumn{1}{|l|}{Qwen2.5-VL \cite{bai2025qwen2} (72B)}      & \multicolumn{1}{c|}{\cmark}      & \multicolumn{1}{c|}{\cmark}      & \multicolumn{1}{c|}{\xmark}  &     \textbf{10.4}        \\ \hline
\multicolumn{1}{|l|}{Qwen2.5-VL \cite{bai2025qwen2} (72B)}      & \multicolumn{1}{c|}{\cmark}      & \multicolumn{1}{c|}{\cmark}      & \multicolumn{1}{c|}{\cmark}  &    10.1         \\ \hline \hline
\multicolumn{1}{|l|}{Qwen3-VL \cite{Qwen3-VL} (8B)}      & \multicolumn{1}{c|}{\cmark}      & \multicolumn{1}{c|}{\xmark}      & \multicolumn{1}{c|}{\xmark}  &       10.4      \\ \hline
\multicolumn{1}{|l|}{Qwen3-VL \cite{Qwen3-VL} (8B)}      & \multicolumn{1}{c|}{\cmark}      & \multicolumn{1}{c|}{\cmark}      & \multicolumn{1}{c|}{\xmark}  &       \textbf{11.4}      \\ \hline
\multicolumn{1}{|l|}{Qwen3-VL \cite{Qwen3-VL} (8B)}      & \multicolumn{1}{c|}{\cmark}      & \multicolumn{1}{c|}{\cmark}      & \multicolumn{1}{c|}{\cmark}  &       7.0      \\ \hline \hline
\multicolumn{1}{|l|}{Qwen3-VL \cite{Qwen3-VL} (30B-A3B)}      & \multicolumn{1}{c|}{\cmark}     & \multicolumn{1}{c|}{\xmark}      & \multicolumn{1}{c|}{\xmark}  &   10.7          \\ \hline
\multicolumn{1}{|l|}{Qwen3-VL \cite{Qwen3-VL} (30B-A3B)}      & \multicolumn{1}{c|}{\cmark}      & \multicolumn{1}{c|}{\cmark}      & \multicolumn{1}{c|}{\xmark}  &    \textbf{11.9}         \\ \hline
\multicolumn{1}{|l|}{Qwen3-VL \cite{Qwen3-VL} (30B-A3B)}      & \multicolumn{1}{c|}{\cmark}      & \multicolumn{1}{c|}{\cmark}      & \multicolumn{1}{c|}{\cmark}  &        9.8     \\ \hline \hline
%\multicolumn{1}{|l|}{Qwen3-VL \cite{Qwen3-VL} (235B-A22B-FP8)}      & \multicolumn{1}{c|}{\cmark}      & \multicolumn{1}{c|}{\xmark}      & \multicolumn{1}{c|}{\xmark}  &       11.7      \\ \hline
%\multicolumn{1}{|l|}{Qwen3-VL \cite{Qwen3-VL} (235B-A22B-FP8)}      & \multicolumn{1}{c|}{\cmark}      & \multicolumn{1}{c|}{\cmark}      & \multicolumn{1}{c|}{\xmark}  &      14.0       \\ \hline
%\multicolumn{1}{|l|}{Qwen3-VL \cite{Qwen3-VL} (235B-A22B-FP8)}      & \multicolumn{1}{c|}{\cmark}      & \multicolumn{1}{c|}{\cmark}      & \multicolumn{1}{c|}{\cmark}  &       9.8      \\ \hline
\multicolumn{1}{|l|}{Gemini 3 Pro \cite{DeepMind2025Gemini3Pro}}      & \multicolumn{1}{c|}{\cmark}      & \multicolumn{1}{c|}{\xmark}      & \multicolumn{1}{c|}{\xmark}  &      21.9       \\ \hline
\multicolumn{1}{|l|}{Gemini 3 Pro \cite{DeepMind2025Gemini3Pro}}      & \multicolumn{1}{c|}{\cmark}      & \multicolumn{1}{c|}{\cmark}      & \multicolumn{1}{c|}{\xmark}  &      23.0       \\ \hline
\multicolumn{1}{|l|}{Gemini 3 Pro \cite{DeepMind2025Gemini3Pro}}      & \multicolumn{1}{c|}{\cmark}      & \multicolumn{1}{c|}{\cmark}      & \multicolumn{1}{c|}{\cmark}  &      \textbf{23.9}     \\ \hline
\end{tabular}
}
\end{table}
}

\begin{figure}[t]
    \centering
    \includegraphics[width=\linewidth]{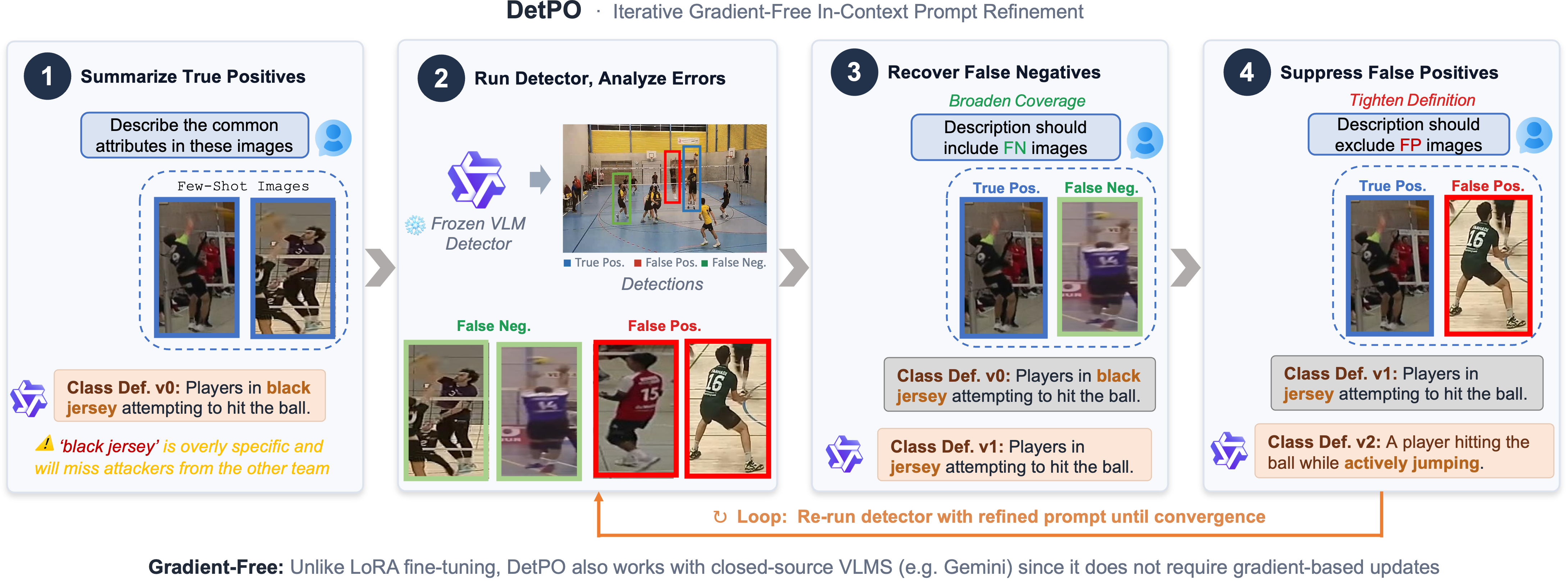}
    \caption{{\bf Detection Prompt Optimization.}
    Our iterative framework begins by generating an initial class definition from a set of few-shot images. The resulting prompt is evaluated on the training set, yielding false positive and false negative predictions. We ask a critique model to use these false negatives to broaden the initial class definition, and false positives to tighten it. We iteratively evaluate and improve the prompt on the training set until convergence. As shown above, the false negative sample removes the constraint ``black jersey'' from the prompt, while the false positive sample adds the attribute ``actively jumping'' to better distinguish the target class from similar actions.
    }
    \label{fig:detpo_method_overview}
\end{figure}

\section{Introduction}
Multi-Modal LLMs (MLLMs) achieve remarkable zero-shot performance across diverse vision-language tasks like image captioning, OCR, and VQA \cite{bai2025qwen2,comanici2025gemini,achiam2023gpt}. On flagship vision tasks like object detection, generalist MLLMs like Qwen3-VL \cite{Qwen3-VL} perform on-par with specialist object detectors like GroundingDINO \cite{liu2024grounding} on popular benchmarks like RefCOCO \cite{yu2016modeling} and OdinW-13 \cite{li2022elevater}. However, such models struggle to generalize to out-of-distribution concepts (e.g. material property estimation, defect detection, and contextual action recognition) and imaging modalities (e.g. X-rays, thermal spectrum data, and aerial imagery) not typically found in internet-scale pre-training \cite{robicheaux2025roboflow100}. We argue that some data will \textit{always} remain out-of-distribution, whether due to being sequestered from the internet or being created after a model's training cutoff. This motivates the need to learn from few-shot examples, similar to how human annotators are taught new concepts via few-shot multi-modal instructions~\cite{madan2024revisiting}. %In such cases, generalist MLLMs will need to learn concepts few-shot examples. 
%While gradient-based few-shot learning of MLLMs %(particularly with parameter-efficient techniques like LORA~\cite{hu2021lora})
%is one approach, this may not be feasible for black-box closed-source models. Morever, even recent open-weights models may be too expensive to fine-tune on consumer-grade hardware. 
 
 \textbf{Few-Shot Detection via In-Context Learning (ICL).}
%Aligning foundation models to target concepts can be framed as a few-shot learning problem, where vision-language models (VLMs) are provided with visual examples and detailed textual descriptions for each class. 
Since frontier MLLMs are typically only accessible via APIs, and state-of-the-art open-weights models are prohibitively expensive to fine-tune on consumer-grade hardware, we argue that gradient-free few-shot learning is best framed through the lens of multi-modal in-context learning (ICL). Although ICL has been extensively studied for LLMs \cite{brown2020language, dong2024survey}, multi-modal ICL for object detection poses unique challenges. We primarily explore this problem using the recently released Roboflow20-VL (RF20-VL) \cite{robicheaux2025roboflow100} benchmark, which aggregates 20 distinct datasets from domains typically not found in internet-scale pre-training. Importantly, object classes are specified via few-shot visual examples and rich textual instructions. In practice, we find that prompting with few-shot visual examples yields inconsistent benefits compared to using class names and instructions (Table \ref{tab:naive_icl}). %Intuitively, more data should help.%multi-modal in-context examples should improve detection accuracy by resolving semantic ambiguities and providing rich contextual cues. However, 
 We posit that this arises from rigid prompt structures used during post-training, making it difficult to exploit additional multi-modal contextual information during zero-shot inference.

 \textbf{Detection Prompt Optimization Improves Concept Alignment.} To address these limitations, we introduce Detection Prompt Optimization (DetPO), a black-box optimization approach that iteratively refines text-only prompts by maximizing detection accuracy on few-shot visual training examples. Unlike classical detectors that primarily learn from true positive examples, we find that MLLMs benefit from seeing \textit{corner cases} and \textit{negative examples}, akin to how human annotators refine their understanding of a target class by learning what \textit{not} to annotate. At each iteration, DetPO refines a text-only prompt based on the model's true positive, false positive, and false negative predictions on the few-shot training set.

 \textbf{DetPO Down-Weights False Positive Detections.} We further observe that current MLLMs often overpredict bounding boxes and lack per-box confidence scores by default. We find that simply prompting models for per-box scores improves detection accuracy without additional compute cost. However, these self-reported confidence scores can sometimes be poorly calibrated. In such cases, we can optionally post-process detections with VQA Score \cite{lin2024evaluating}. We prompt the model with bounding box predictions overlaid on the test image and ask ``Is this bounding box an instance of class {\tt \{CLS\}}?'' We use the normalized probability of the {\tt yes} token as the final bounding box confidence score. This simple post-processing step down-weights false positives and improves mAP. 

 \textbf{Contributions.} We present three major contributions. First, we benchmark state-of-the-art MLLMs on RF20-VL and LVIS and show that naively prompting with multi-modal in-context examples yields poor performance. Next, we propose Detection Prompt Optimization (DetPO) to iteratively use model predictions on the training set to refine text-only prompts and estimate better per-box confidence scores to improve concept alignment. Lastly, we extensively ablate our design choices and demonstrate that DetPO consistently improves performance across popular MLLMs, establishing a new state-of-art for black-box few-shot object detection. %61.3\%. 

\section{Related Works}

 \textbf{Vision-Language Models} are often pre-trained on large-scale, weakly supervised image-text pairs \cite{schuhmann2022laion} before fine-tuning on task-specific data. While early VLMs mainly focused on image classification \cite{radford2021learning} and VQA, recent methods have extended these models for visual grounding through open-vocabulary detection. Early efforts adapted VLMs for detection by classifying region proposals \cite{gu2021vild, gu2021open} or by integrating detection heads into either frozen \cite{kuo2022fvlm} or fine-tuned \cite{minderer2022owlvit, minderer2023owlvit2, du2022learning} encoders. RegionCLIP \cite{zhong2022regionclip} introduced a multi-stage pipeline that combined pseudo-label generation, region-text contrastive pre-training, and fine-tuning on detection benchmarks. GLIP \cite{li2021grounded} reformulates detection as a phrase grounding task, where a single text query is applied to the entire image. Detic \cite{zhou2022detecting} improves long-tail detection performance by leveraging image-level supervision from ImageNet's 21K classes \cite{russakovsky2015imagenet}. Modern VLMs demonstrate strong zero-shot performance and are often employed as ``black-box'' tools in downstream tasks \cite{ma2023long, pmlr-v205-peri23a, khurana2024shelf, sal2024eccv, takmaz2025cal}. More recently, multi-modal large language models (MLLMs) such as Qwen2.5-VL \cite{bai2025qwen2}, Qwen3-VL \cite{Qwen3-VL}, Gemini 2.5 Pro \cite{google_gemini_2024}, and Gemini 3 Pro \cite{DeepMind2025Gemini3Pro} reframe spatial understanding as a next-token prediction task. Notably, generalist MLLMs demonstrate strong zero-shot visual grounding capabilities on-par with specialist models like GroundingDINO \cite{liu2023grounding} on popular object detection benchmarks like OdinW-13 \cite{li2022elevater} and RefCOCO \cite{yu2016modeling}. However, we find that these generalist models achieve poor zero-shot detection accuracy on out-of-distribution classes, tasks, and modalities, motivating the need for few-shot concept alignment. 

 \textbf{Few-Shot Object Detection (FSOD)} focuses on recognizing novel object categories with limited training examples \cite{kohler2021few}. Prior work primarily explores two paradigms: meta-learning and transfer learning. Meta-learning methods aim to learn transferable representations from {\tt base} classes that can generalize to unseen {\tt novel} classes. For instance, Kang et al. \cite{kang2019few} re-weights features from {\tt base} classes to infer {\tt novel} ones, while Xiao et al. \cite{xiao2022few} unifies few-shot detection and viewpoint estimation. Fan et al. \cite{fan2020few} develops a matching-based few-shot object detection network that learns a similarity metric between image pairs, and Wu et al. \cite{wu2021universal} enhances object features using universal prototypes. Xu et al. \cite{xu2023generating} further propose a generative framework that improves robustness to noisy object proposals. In contrast, transfer learning methods partially freeze weights pretrained on {\tt base} datasets to enable adaptation to {\tt novel} classes with limited samples. These methods typically adopt a two-stage fine-tuning process, training on {\tt base} classes followed by refinement of the box classifier and regressor using $K$-shot examples. This strategy has generally outperformed meta-learning approaches \cite{wang2020frustratingly}. Different from prior work, we explore few-shot object detection for generalist MLLMs using multi-modal in-context learning.

 \textbf{In-Context Learning (ICL)} is an emergent capability that enables LLMs to reason by analogy \cite{dong2024survey}. Brown et al. \cite{brown2020language} first popularized few-shot learning via ICL. Subsequent methods extend this idea by improving reasoning through structured prompting, such as decomposing complex instructions into simpler steps \cite{wei2022chain, yao2023tree}. Recently, ICL has also gained traction in the vision-language community. Flamingo \cite{alayrac2022flamingo} demonstrated that large-scale VLMs can perform ICL effectively, improving tasks like image captioning with only a few examples. More recently, Emu 2 \cite{sun2024generative} demonstrated that scaling encoder-decoder architectures with auto-regressive training improves ICL. Different from prior work, we address multi-modal in-context learning of object detection.

 \textbf{Prompt Optimization} seeks to automate prompt engineering to maximize model performance on downstream tasks \cite{ramnath2025systematic}. This problem has been extensively studied for LLMs, where prompting provides a natural and flexible interface for humans to interact with generalist models. Notably, prompting has become a standard approach for solving many flagship NLP tasks \cite{schick2021exploiting, brown2020language, sanh2021multitask}. However, since LLMs are particularly sensitive to user prompts \cite{webson2022prompt}, they often require careful prompt design \cite{reynolds2021prompt, shin2020autoprompt}. Soft prompt tuning methods demonstrate strong performance using gradient-based optimization \cite{liu2024gpt, qin2021learning, lester2021power}. However, such optimization techniques are impractical for frontier MLLMs that typically only accessible via APIs, and  state-of-the-art open-weights models are prohibitively expensive to fine-tune on consumer-grade hardware. In this work, we draw inspiration from discrete prompt search methods such as prompt generation \cite{ben2022pada, agrawal2025gepa, opsahl2024optimizing}, prompt scoring \cite{feldman2019commonsense}, and prompt paraphrasing \cite{jiang2020can, yuan2021bartscore} to optimize class descriptions directly in natural language space.

\section{Detection Prompt Optimization for FSOD}
In this section, we present our Detection Prompt Optimization (DetPO) framework for improving MLLM alignment to target concepts using few-shot multi-modal examples. Existing prompt optimizers treat the downstream (visual) task as a black-box to be ``blindly" optimized via a numeric reward. For example, GEPA \cite{agrawal2025gepa} refines a set of candidate prompts using only the numeric IoU's of predicted bounding boxes compared to the ground-truth. In contrast, DetPO directly feeds visual examples of false positives and false negatives to an MLLM to generate the refined class prompt; this allows, for example, the MLLM to ``see'' that occluded ground-truth objects are being systematically missed and update the prompt accordingly. Our key insight is that {\em visual tasks should leverage visual feedback during prompt optimization}. We describe our algorithm below and provide psuedo-code in Appendix \ref{sec:algo}. 

\begin{figure}[t]
\centering
\includegraphics[width=\linewidth]{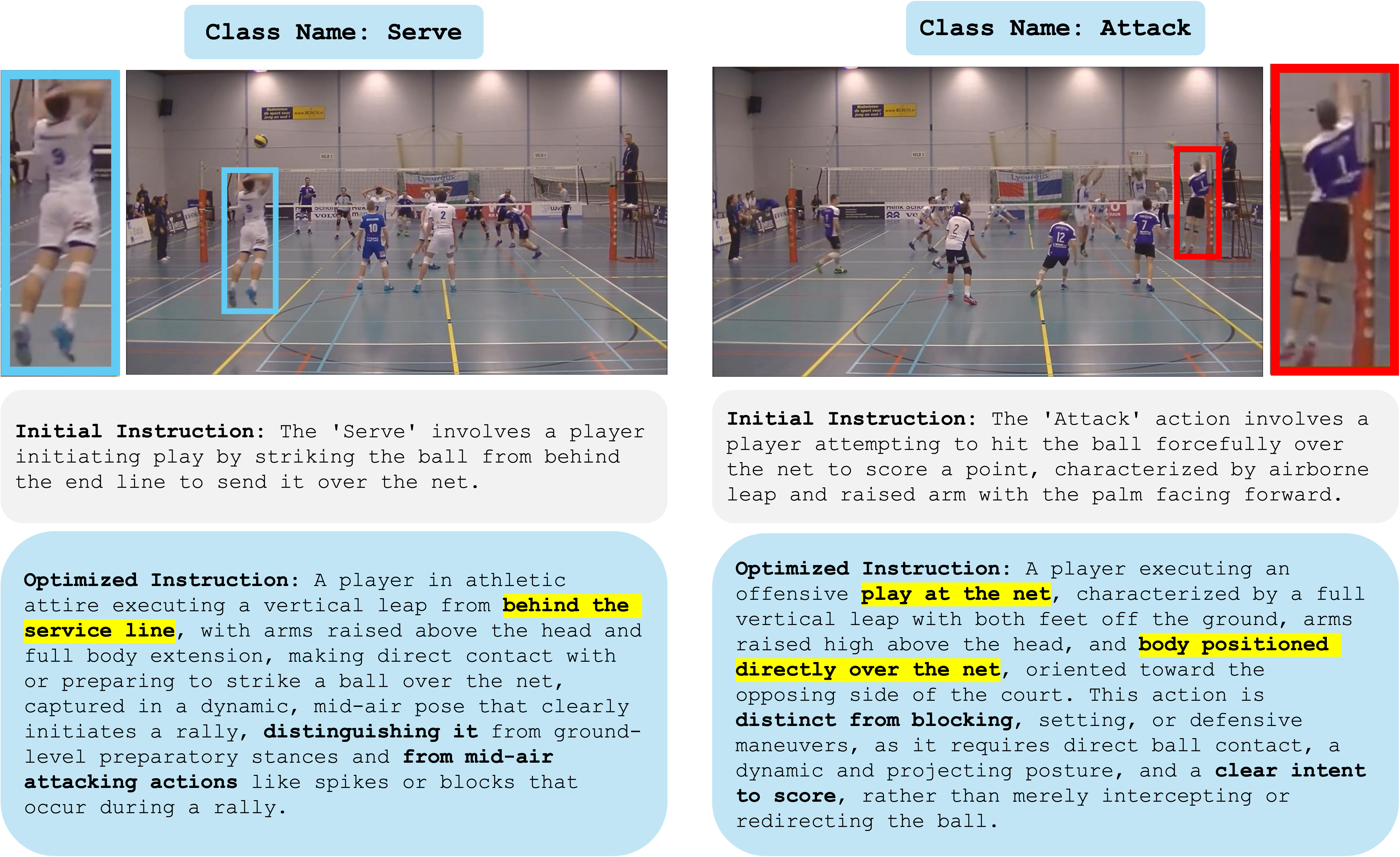}
    \caption{{\bf Contrastive Prompt Refinement Reduces Class Confusion.} We present an example of prompt refinement on the challenging Actions dataset above. Even for the visually similar {\tt Attack} and {\tt Serve} classes, which both depict a player airborne while striking the ball, our method discovers discriminative attributes that reduce class confusion.  Specifically, DetPO is able to differentiate that {\tt Serve} occurs behind the service line, while {\tt Attack} is played at the net.
    }
    \label{fig:contrastive_prompt}
\end{figure}

\textbf{Constrastive Prompt Refinement.} Unlike traditional detectors, which are trained primarily on true positive examples, we find that MLLMs achieve higher detection accuracy when provided with corner cases and negative examples, similar to how human annotators learn from examples of what \textit{not} to annotate. Figure~\ref{fig:detpo_method_overview} illustrates the prompt refinement procedure of our Detection Prompt Optimization (DetPO) framework. For each class $C$, we generate an initial class description by prompting the model to describe key features of all ground-truth instances of $C$ in the training set. We then refine this description by prompting the model to contrast $C$ with ground-truth instances from all other classes, encouraging the model to focus on discriminative features. 

\begin{figure}[t]
  \centering
  \begin{subfigure}[t]{0.49\columnwidth}
    \includegraphics[width=\linewidth]{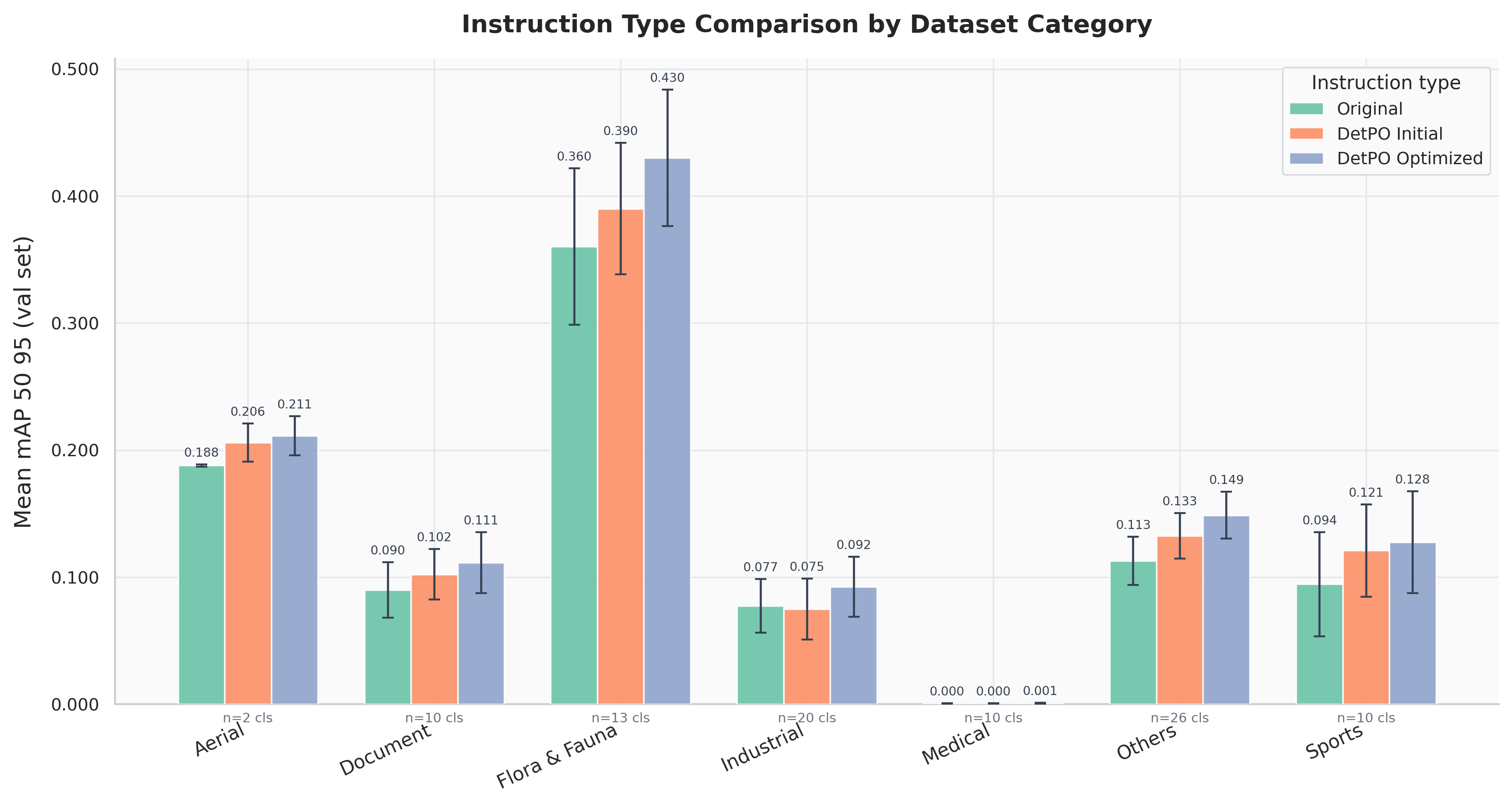}
    \caption{Instruction Type Comparision}
    \end{subfigure}
  \hfill
    \begin{subfigure}[t]{0.49\columnwidth}
    \includegraphics[width=\linewidth]{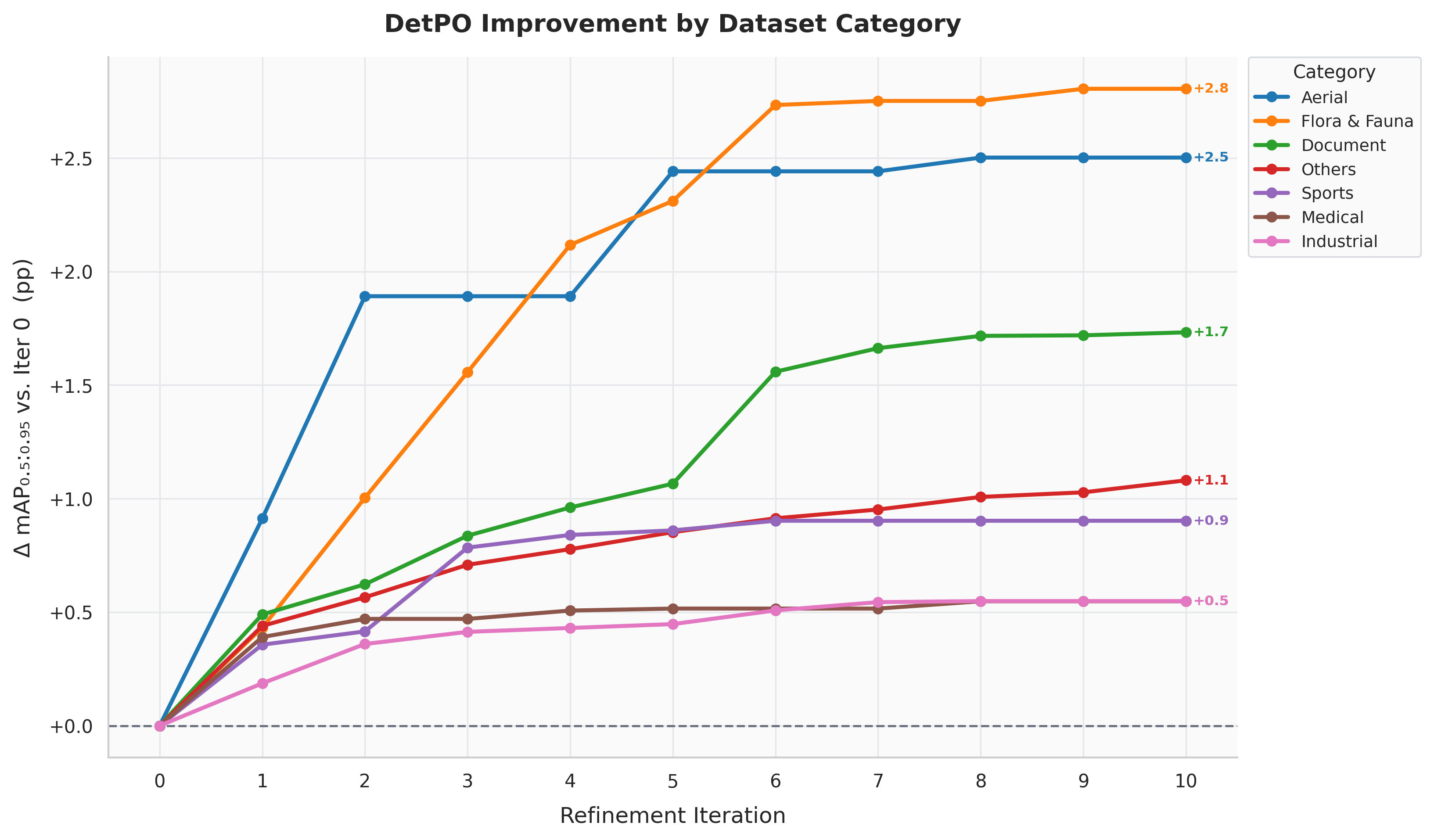}
    \caption{Iterative Accuracy Improvement}
  \end{subfigure}
  \caption{\textbf{Improvement from Contrastive Prompt Refinement.} We compare the original baseline prompt against both the initial DetPO prompt and the final optimized DetPO prompt (\textbf{left}). These results demonstrate that the DetPO optimized prompts consistently improves detection accuracy across nearly all categories. Further, we show that successive refinement iterations improves performance on the training set (\textbf{right}). Importantly, we plot the change in mAP relative to the initial DetPO prompt. Most domains show strong initial gains that begin to plateau around iteration 6, with the Flora \& Fauna and Aerial categories showing the largest overall improvements (+2.8 and +2.5, respectively).}
\label{fig:detpo_plot}
\end{figure}

Using this detailed initial prompt, we run inference on the training images to generate candidate detections and identify all false positives and false negatives. Next, we find the top $K$ most severe false positives and false negatives (e.g., the false positives with the highest confidence scores and the false negatives with the lowest IoU), and use these predictions to guide the next refinement step (Figure \ref{fig:detpo_confusion}). For the sampled false positive, we instruct the critique model to revise the class definition to explicitly \textit{exclude} the incorrect instance, whereas for the sampled false negative, we prompt the critique model to revise the definition to explicitly \textit{include} the missed instance. To guide this revision, the critique model is first asked to identify the key differences between a reference ground-truth instance and the false positive, or the key similarities between the reference instance and the false negative, and then update the class definition accordingly. To better capture contextual cues, we prompt the model with full images, highlighting the reference ground-truth instance and the false positive or false negative using different-colored bounding boxes. After updating the prompt, we rerun inference on the few-shot training set and repeat this process until training set performance converges or a maximum number of refinement steps is reached (Figure \ref{fig:detpo_plot}). 

Finally, we generate multiple candidate refinements of the optimized prompt and evaluate them, together with the initial prompt, on the few-shot validation set. The prompt achieving the highest validation performance is selected for inference on the test set.

\textbf{Confidence Score Estimation.} Although specialist detectors typically predict per-box confidence scores, MLLMs do not by default. We find that simply prompting the model to predict a confidence score per-box works well in practice, down weighting false positives more effectively than the baseline prompt. We include all prompt templates in Appendix \ref{sec:prompts}.

\textbf{VQA Score.} Directly asking the model to self-report bounding box confidence scores inside the optimization loop effectively balances iteration speed and confidence estimation accuracy. However, it does not explicitly incorporate object-specific visual features. To address this limitation, we post-process the confidence scores after optimization for each predicted bounding box on the test set using VQA Score \cite{lin2024evaluating}. Specifically, for each prediction, we draw a box on the original image and prompt the model with the binary question: ``Is there an instance of class {\tt \{CLS\}} inside this bounding box? Please answer Yes or No.'' We compute the final confidence score as the normalized likelihood of the \texttt{Yes} token (i.e., $\frac{p(\texttt{Yes})}{p(\texttt{Yes}) + p(\texttt{No})}$). Although VQA Score provides higher quality confidence estimates than self-reported scores, its runtime scales linearly with the number of predictions, making it computationally expensive in practice. We include this optional re-ranking step as an upper bound on the performance of our approach, but note that using self-reported confidence scores is a strong alternative (Table \ref{tab:transfer}). Notably, recent black-box LLMs like Gemini 3 Pro \cite{DeepMind2025Gemini3Pro} do not expose token probabilities in their API, likely to hamper efforts to distill the model's predictions. Therefore, we utilize Qwen3-VL (30B-A3B) to post-process Gemini's predictions to generate VQA-based confidence scores. Our results show that DetPO is robust to ensembling different black-box models.

\section{Experiments}
In this section, we evaluate DetPO against recent state-of-the-art specialist and generalist detectors. Further, we ablate our design choices to highlight the impact of contrastive prompt refinement and confidence score estimation.

\textbf{Datasets and Metrics.}
We evaluate DetPO on Roboflow20-VL (RF20-VL), a subset of the Roboflow100-VL \cite{robicheaux2025roboflow100} benchmark. RF20-VL is a few-shot object detection benchmark curated from Roboflow Universe, a community-driven platform that hosts diverse open-source datasets for real-world computer vision tasks. Notably, the benchmark includes 20 datasets spanning diverse domains such as aerial imagery, X-rays, medical imaging, and wildlife monitoring. Each dataset provides 10-shot training examples and rich annotation instructions per class. We follow the standard COCO evaluation protocol \cite{lin2014microsoft} and report mean Average Precision (mAP) for each super-category, as defined by Robicheaux et al. \cite{robicheaux2025roboflow100}, along with the average mAP across all 20 datasets. We refer readers to Appendix \ref{sec:implementation_details} for full details of our experimental setup, model configurations, and prompts, and Appendix \ref{sec:lvis} for benchmarking results on LVIS.

\textbf{Baselines.} We evaluate the zero-shot performance of specialist detectors like GroundingDINO \cite{liu2024grounding}, LLMDet \cite{fu2025llmdet}, SAM3 \cite{carion2025sam}, MQ-GLIP \cite{xu2023multi}, and YOLO-E \cite{wang2025yoloerealtimeseeing}, as well as generalist models like Qwen 2.5-VL \cite{bai2025qwen2}, Qwen 3-VL \cite{Qwen3-VL}, and Gemini 3 Pro \cite{DeepMind2025Gemini3Pro}. Further, we evaluate prompt optimization approaches like GEPA \cite{agrawal2025gepa} and MIPROv2 \cite{opsahl2024optimizing} with Qwen 3-VL and Gemini 3 Pro using DSPy \cite{khattab2023dspy}.

{
\setlength{\tabcolsep}{0.2em}
\begin{table*}[t]
\centering
\caption{\textbf{Roboflow20-VL Benchmark.} We evaluate recent specialist and generalist models on 20 datasets from the Roboflow20-VL (RF20-VL) benchmark. Notably, we find that specialist object detectors such as MQ-GLIP and YOLO-E fail to benefit from few-shot visual examples, with performance capped at 14.0 mAP. The best specialist model, LLMDet (17.2 mAP), relies solely on class name inputs. In contrast, generalist MLLMs like Qwen3-VL (30B-A3B) and Gemini 3 Pro augmented with DetPO prompts outperform specialist detectors, reaching 21.6 and 26.4 mAP respectively. Across all evaluated generalist models, our Detection Prompt Optimization (DetPO) framework substantially improves accuracy over the baseline. Note that C is class names, I is instructions, and V is for images.
} 
\label{tab:rf20-vl}
\scalebox{0.70}{
\begin{tabular}{|lcccccccc|}
\hline

\rowcolor{gray!10}
\multicolumn{1}{|l|}{\textbf{Method}}                     & \multicolumn{1}{c|}{\textbf{Aerial}} & \multicolumn{1}{c|}{\textbf{Document}} & \multicolumn{1}{c|}{\textbf{Flora \& Fauna}} & \multicolumn{1}{c|}{\textbf{Industrial}} & \multicolumn{1}{c|}{\textbf{Medical}} & \multicolumn{1}{c|}{\textbf{Sports}} & \multicolumn{1}{c|}{\textbf{Other}} & \textbf{All} \\ \hline
\rowcolor{gray!10}
\multicolumn{9}{|l|}{\textbf{Specialist Models}} \\ \hline                                            
\multicolumn{1}{|l|}{GroundingDINO \cite{liu2024grounding} (C)}                       & \multicolumn{1}{c|}{28.5}      & \multicolumn{1}{c|}{5.1}      & \multicolumn{1}{c|}{33.7}         & \multicolumn{1}{c|}{12.8}                    & \multicolumn{1}{c|}{0.4}                 & \multicolumn{1}{c|}{5.1}                & \multicolumn{1}{c|}{16.9}               &       16.8       \\ \hline
\multicolumn{1}{|l|}{LLMDet \cite{fu2025llmdet} (C)}                       & \multicolumn{1}{c|}{32.3}      & \multicolumn{1}{c|}{4.4}      & \multicolumn{1}{c|}{33.6}         & \multicolumn{1}{c|}{12.6}                    & \multicolumn{1}{c|}{0.7}                 & \multicolumn{1}{c|}{6.7}                & \multicolumn{1}{c|}{16.7}               &       \textbf{17.2}       \\ \hline
\multicolumn{1}{|l|}{SAM3 \cite{carion2025sam} (C)}                       & \multicolumn{1}{c|}{32.3}      & \multicolumn{1}{c|}{15.3}      & \multicolumn{1}{c|}{17.1}         & \multicolumn{1}{c|}{13.8}                    & \multicolumn{1}{c|}{2.0}                 & \multicolumn{1}{c|}{14.2}                & \multicolumn{1}{c|}{17.8}               &       16.3       \\ \hline \hline
\multicolumn{1}{|l|}{MQ-GLIP \cite{xu2023multi} (C)}     & \multicolumn{1}{c|}{30.1}      & \multicolumn{1}{c|}{2.5}      & \multicolumn{1}{c|}{32.8}         & \multicolumn{1}{c|}{5.5}                    & \multicolumn{1}{c|}{0.5}                 & \multicolumn{1}{c|}{6.4}                & \multicolumn{1}{c|}{10.8}               &      14.0       \\ \hline
\multicolumn{1}{|l|}{MQ-GLIP \cite{xu2023multi} (V)}         & \multicolumn{1}{c|}{1.8}      & \multicolumn{1}{c|}{1.1}      & \multicolumn{1}{c|}{17.6}         & \multicolumn{1}{c|}{1.8}                    & \multicolumn{1}{c|}{0.1}                 & \multicolumn{1}{c|}{6.6}                & \multicolumn{1}{c|}{6.8}               &      6.7        \\ \hline
\multicolumn{1}{|l|}{MQ-GLIP \cite{xu2023multi} (C + V)}      & \multicolumn{1}{c|}{29.8}      & \multicolumn{1}{c|}{2.5}      & \multicolumn{1}{c|}{32.7}         & \multicolumn{1}{c|}{5.6}                    & \multicolumn{1}{c|}{0.5}                 & \multicolumn{1}{c|}{6.5}                & \multicolumn{1}{c|}{10.9}               &       14.0       \\ \hline \hline
\multicolumn{1}{|l|}{YOLO-E \cite{wang2025yoloerealtimeseeing} (C}      & \multicolumn{1}{c|}{10.2}      & \multicolumn{1}{c|}{1.6}      & \multicolumn{1}{c|}{16.4}         & \multicolumn{1}{c|}{8.1}                    & \multicolumn{1}{c|}{0.3}                 & \multicolumn{1}{c|}{7.8}                & \multicolumn{1}{c|}{10.9}               &      9.2       \\ \hline
\multicolumn{1}{|l|}{YOLO-E \cite{wang2025yoloerealtimeseeing} (V)}      & \multicolumn{1}{c|}{11.6}      & \multicolumn{1}{c|}{10.7}      & \multicolumn{1}{c|}{17.5}         & \multicolumn{1}{c|}{15.2}                    & \multicolumn{1}{c|}{2.3}                 & \multicolumn{1}{c|}{8.5}                &  \multicolumn{1}{c|}{16.4}               &       13.2      \\ \hline
\multicolumn{1}{|l|}{YOLO-E \cite{wang2025yoloerealtimeseeing} (C + V)}      & \multicolumn{1}{c|}{12.8}      & \multicolumn{1}{c|}{6.1}      & \multicolumn{1}{c|}{21.0}         & \multicolumn{1}{c|}{14.7}                    & \multicolumn{1}{c|}{1.7}                 & \multicolumn{1}{c|}{10.9}                & \multicolumn{1}{c|}{15.7}               &       13.4      \\ \hline \hline
\rowcolor{gray!10}
\multicolumn{9}{|l|}{\textbf{Generalist Models}}  \\  \hline 
\multicolumn{1}{|l|}{Qwen3-VL \cite{Qwen3-VL} (30B-A3B) (C + I)}     & \multicolumn{1}{c|}{9.0}      & \multicolumn{1}{c|}{7.8}      & \multicolumn{1}{c|}{23.5}         & \multicolumn{1}{c|}{9.6}                    & \multicolumn{1}{c|}{0.7}                 & \multicolumn{1}{c|}{14.4}                & \multicolumn{1}{c|}{10.1}               &      11.9        \\ \hline
\multicolumn{1}{|l|}{\quad w/ GEPA \cite{agrawal2025gepa}}      & \multicolumn{1}{c|}{9.3}      & \multicolumn{1}{c|}{12.4}      & \multicolumn{1}{c|}{23.6}         & \multicolumn{1}{c|}{10.8}                    & \multicolumn{1}{c|}{1.3}                 & \multicolumn{1}{c|}{15.1}                & \multicolumn{1}{c|}{11.3}               &     13.0        \\ \hline
% \multicolumn{1}{|l|}{\quad \quad + GPT 5 Reflection LLM}      & \multicolumn{1}{c|}{}      & \multicolumn{1}{c|}{}      & \multicolumn{1}{c|}{}         & \multicolumn{1}{c|}{}                    & \multicolumn{1}{c|}{}                 & \multicolumn{1}{c|}{}                & \multicolumn{1}{c|}{}               &             \\ \hline
\multicolumn{1}{|l|}{\quad w/ MIPROv2 \cite{opsahl2024optimizing}}      & \multicolumn{1}{c|}{8.7}      & \multicolumn{1}{c|}{5.6}      & \multicolumn{1}{c|}{18.6}         & \multicolumn{1}{c|}{10.3}                    & \multicolumn{1}{c|}{0.0}                 & \multicolumn{1}{c|}{15.1}                & \multicolumn{1}{c|}{9.9}               &     10.7        \\ \hline
%\multicolumn{1}{|l|}{\quad w/ DetPO (Ours)}      & \multicolumn{1}{c|}{13.8}      & \multicolumn{1}{c|}{18.6}      & \multicolumn{1}{c|}{34.6}         & \multicolumn{1}{c|}{19.7}                    & \multicolumn{1}{c|}{0.1}                 & \multicolumn{1}{c|}{21.8}                & \multicolumn{1}{c|}{16.4}               &     19.4         \\ \hline
\multicolumn{1}{|l|}{\quad w/ DetPO (Ours) + VQA Score \cite{lin2024evaluating}}      & \multicolumn{1}{c|}{16.1}      & \multicolumn{1}{c|}{25.2}      & \multicolumn{1}{c|}{36.5}         & \multicolumn{1}{c|}{20.1}                    & \multicolumn{1}{c|}{0.2}                 & \multicolumn{1}{c|}{25.7}                & \multicolumn{1}{c|}{18.4}               &     \textbf{21.6}         \\ \hline \hline
\multicolumn{1}{|l|}{Gemini 3 Pro \cite{DeepMind2025Gemini3Pro} (C + I + V)}      & \multicolumn{1}{c|}{27.0}      & \multicolumn{1}{c|}{26.7}      & \multicolumn{1}{c|}{31.3}         & \multicolumn{1}{c|}{26.2}                    & \multicolumn{1}{c|}{2.6}                 & \multicolumn{1}{c|}{26.9}                & \multicolumn{1}{c|}{13.3}               &       23.8       \\ \hline
\multicolumn{1}{|l|}{\quad w/ GEPA \cite{agrawal2025gepa}}      & \multicolumn{1}{c|}{19.2}      & \multicolumn{1}{c|}{30.6}      & \multicolumn{1}{c|}{32.1}         & \multicolumn{1}{c|}{32.7}                    & \multicolumn{1}{c|}{2.0}                 & \multicolumn{1}{c|}{28.2}                & \multicolumn{1}{c|}{20.8}               &     25.6        \\ \hline
\multicolumn{1}{|l|}{\quad w/ MIPROv2 \cite{opsahl2024optimizing}}      & \multicolumn{1}{c|}{22.9}      & \multicolumn{1}{c|}{27.0}      & \multicolumn{1}{c|}{31.6}         & \multicolumn{1}{c|}{26.4}                    & \multicolumn{1}{c|}{3.0}                 & \multicolumn{1}{c|}{29.7}                & \multicolumn{1}{c|}{19.8}               &        25.0     \\ \hline
%\multicolumn{1}{|l|}{\quad w/ DetPO$^*$ (Ours)}      & \multicolumn{1}{c|}{27.0}      & \multicolumn{1}{c|}{35.4}      & \multicolumn{1}{c|}{35.7}         & \multicolumn{1}{c|}{23.0}                    & \multicolumn{1}{c|}{3.8}                 & \multicolumn{1}{c|}{28.5}                & \multicolumn{1}{c|}{20.8}               &       26.4        \\ \hline
\multicolumn{1}{|l|}{\quad w/ DetPO (Ours) + VQA Score \cite{lin2024evaluating}}      & \multicolumn{1}{c|}{26.2}      & \multicolumn{1}{c|}{35.7}      & \multicolumn{1}{c|}{35.4}         & \multicolumn{1}{c|}{23.3}                    & \multicolumn{1}{c|}{3.9}                 & \multicolumn{1}{c|}{28.2}                & \multicolumn{1}{c|}{20.4}               &       \textbf{26.3}      \\ \hline
\end{tabular}
}
\end{table*}
}

{
\setlength{\tabcolsep}{1.2em}
\begin{table}[t]
\centering
\caption{\textbf{DetPO Transfers Across Generalist Detectors}. We evaluate our detection prompt optimization approach across popular MLLMs like Qwen 2.5-VL and Qwen 3-VL. Notably, we find that DetPO consistently improves over the baseline. Further, VQA Score yields modest improvements for all models, highlighting the importance of well-calibrated confidence score estimates. 
} 
\label{tab:transfer}
\scalebox{0.70}{
\begin{tabular}{|lcccccccc|}
\hline

\rowcolor{gray!10}
\multicolumn{1}{|l|}{\textbf{Method}}                     & \multicolumn{1}{c|}{\textbf{A}} & \multicolumn{1}{c|}{\textbf{D}} & \multicolumn{1}{c|}{\textbf{F \& F}} & \multicolumn{1}{c|}{\textbf{I}} & \multicolumn{1}{c|}{\textbf{M}} & \multicolumn{1}{c|}{\textbf{S}} & \multicolumn{1}{c|}{\textbf{O}} & \textbf{All} \\ \hline
\multicolumn{1}{|l|}{Qwen2.5-VL (7B) \cite{bai2025qwen2} (C + I)}      & \multicolumn{1}{c|}{4.9}      & \multicolumn{1}{c|}{5.1}      & \multicolumn{1}{c|}{13.5}         & \multicolumn{1}{c|}{3.9}                    & \multicolumn{1}{c|}{0.1}                 & \multicolumn{1}{c|}{7.3}                & \multicolumn{1}{c|}{5.0}               &      6.2        \\ \hline
\multicolumn{1}{|l|}{\quad w/ DetPO (Ours)}      & \multicolumn{1}{c|}{6.2}      & \multicolumn{1}{c|}{12.1}      & \multicolumn{1}{c|}{19.3}         & \multicolumn{1}{c|}{4.2}                    & \multicolumn{1}{c|}{0.0}                 & \multicolumn{1}{c|}{9.1}                & \multicolumn{1}{c|}{7.5}               &     9.1         \\ \hline 
\multicolumn{1}{|l|}{\quad \quad + VQA Score \cite{lin2024evaluating}}      & \multicolumn{1}{c|}{9.4}      & \multicolumn{1}{c|}{17.3}      & \multicolumn{1}{c|}{23.4}         & \multicolumn{1}{c|}{7.2}                    & \multicolumn{1}{c|}{0.0}                 & \multicolumn{1}{c|}{12.8}                & \multicolumn{1}{c|}{8.8}               &     \textbf{11.9}        \\ \hline \hline
\multicolumn{1}{|l|}{Qwen2.5-VL (72B) \cite{bai2025qwen2} (C + I)}      & \multicolumn{1}{c|}{6.3}      & \multicolumn{1}{c|}{10.7}      & \multicolumn{1}{c|}{19.0}         & \multicolumn{1}{c|}{7.5}                    & \multicolumn{1}{c|}{0.4}                 & \multicolumn{1}{c|}{14.4}                & \multicolumn{1}{c|}{9.1}               &      10.4        \\ \hline
\multicolumn{1}{|l|}{\quad w/ DetPO (Ours)}      & \multicolumn{1}{c|}{11.1}      & \multicolumn{1}{c|}{23.0}      & \multicolumn{1}{c|}{26.1}         & \multicolumn{1}{c|}{12.4}                    & \multicolumn{1}{c|}{0.5}                 & \multicolumn{1}{c|}{14.8}                & \multicolumn{1}{c|}{14.9}               &    15.7     \\ \hline 
\multicolumn{1}{|l|}{\quad \quad + VQA Score \cite{lin2024evaluating}}      & \multicolumn{1}{c|}{10.8}      & \multicolumn{1}{c|}{26.3}      & \multicolumn{1}{c|}{26.7}         & \multicolumn{1}{c|}{13.0}                    & \multicolumn{1}{c|}{0.5}                 & \multicolumn{1}{c|}{16.7}                & \multicolumn{1}{c|}{15.0}               &    \textbf{16.5}         \\ \hline \hline
\multicolumn{1}{|l|}{Qwen3-VL \cite{Qwen3-VL} (8B)  (C + I)}      & \multicolumn{1}{c|}{7.1}      & \multicolumn{1}{c|}{7.3}      & \multicolumn{1}{c|}{24.8}         & \multicolumn{1}{c|}{9.3}                    & \multicolumn{1}{c|}{0.2}                 & \multicolumn{1}{c|}{10.2}                & \multicolumn{1}{c|}{10.9}               &      11.4        \\ \hline
\multicolumn{1}{|l|}{\quad w/ DetPO (Ours)}      & \multicolumn{1}{c|}{8.3}      & \multicolumn{1}{c|}{19.1}      & \multicolumn{1}{c|}{30.3}         & \multicolumn{1}{c|}{13.7}                    & \multicolumn{1}{c|}{0.1}                 & \multicolumn{1}{c|}{14.2}                & \multicolumn{1}{c|}{12.2}               &       15.3       \\ \hline 
\multicolumn{1}{|l|}{\quad \quad + VQA Score \cite{lin2024evaluating}}      & \multicolumn{1}{c|}{12.3}      & \multicolumn{1}{c|}{24.2}      & \multicolumn{1}{c|}{32.3}         & \multicolumn{1}{c|}{13.5}                    & \multicolumn{1}{c|}{0.2}                 & \multicolumn{1}{c|}{17.2}                & \multicolumn{1}{c|}{14.3}               &    \textbf{17.5}         \\ \hline \hline
\multicolumn{1}{|l|}{Qwen3-VL \cite{Qwen3-VL} (30B-A3B) (C + I)}      & \multicolumn{1}{c|}{9.0}      & \multicolumn{1}{c|}{7.8}      & \multicolumn{1}{c|}{23.5}         & \multicolumn{1}{c|}{9.6}                    & \multicolumn{1}{c|}{0.7}                 & \multicolumn{1}{c|}{14.4}                & \multicolumn{1}{c|}{10.1}               &      11.9        \\ \hline
\multicolumn{1}{|l|}{\quad w/ DetPO (Ours)}      & \multicolumn{1}{c|}{13.8}      & \multicolumn{1}{c|}{18.6}      & \multicolumn{1}{c|}{34.6}         & \multicolumn{1}{c|}{19.7}                    & \multicolumn{1}{c|}{0.1}                 & \multicolumn{1}{c|}{21.8}                & \multicolumn{1}{c|}{16.4}               &     19.4         \\ \hline
\multicolumn{1}{|l|}{\quad \quad + VQA Score \cite{lin2024evaluating}}      & \multicolumn{1}{c|}{16.1}      & \multicolumn{1}{c|}{25.2}      & \multicolumn{1}{c|}{36.5}         & \multicolumn{1}{c|}{20.1}                    & \multicolumn{1}{c|}{0.2}                 & \multicolumn{1}{c|}{25.7}                & \multicolumn{1}{c|}{18.4}               &     \textbf{21.6}         \\ \hline 
%\multicolumn{1}{|l|}{Gemini 3 Pro \cite{DeepMind2025Gemini3Pro} (C + I + V)}      & \multicolumn{1}{c|}{27.0}      & \multicolumn{1}{c|}{26.7}      & \multicolumn{1}{c|}{31.3}         & \multicolumn{1}{c|}{26.2}                    & \multicolumn{1}{c|}{2.6}                 & \multicolumn{1}{c|}{26.9}                & \multicolumn{1}{c|}{13.3}               &       23.8       \\ \hline
%\multicolumn{1}{|l|}{\quad w/ DetPO$^*$ (Ours)}      & \multicolumn{1}{c|}{27.0}      & \multicolumn{1}{c|}{35.4}      & \multicolumn{1}{c|}{35.7}         & \multicolumn{1}{c|}{23.0}                    & \multicolumn{1}{c|}{3.8}                 & \multicolumn{1}{c|}{28.5}                & \multicolumn{1}{c|}{20.8}               &       26.4        \\ \hline
%\multicolumn{1}{|l|}{\quad \quad + VQA Score}      & \multicolumn{1}{c|}{26.2}      & \multicolumn{1}{c|}{35.7}      & \multicolumn{1}{c|}{35.4}         & \multicolumn{1}{c|}{23.3}                    & \multicolumn{1}{c|}{3.9}                 & \multicolumn{1}{c|}{28.2}                & \multicolumn{1}{c|}{20.4}               &       26.3      \\ \hline
\end{tabular}
}
\end{table}
}

\textbf{DetPO Outperforms Specialist Zero-Shot Models.} While specialist detectors outperform baseline generalist models (with the exception of Gemini 3 Pro) on RF20-VL, prompting generalist models with DetPO's black-box optimized prompts significantly outperforms specialist object detectors (Table \ref{tab:rf20-vl}).
%As shown in Table \ref{tab:rf20-vl}, specialist detectors outperform our baseline generalist detectors (with the execption of Gemini 3 Pro) on RF20-VL. However, we find that prompting generalist models with DetPO optimized prompts significantly improves performance, outperforming recent specialist object detectors on the Roboflow20-VL benchmark. 
While top-performing specialist models like LLMDet and GroundingDINO achieve 17.2 and 16.8 AP, respectively, their test-time performance is fundamentally limited because it is difficult to effectively leverage few-shot visual examples. In contrast, applying DetPO to generalist models yields substantial gains. Gemini 3 Pro with DetPO prompts achieves a state-of-the-art 26.4 mAP, outperforming the best specialist model by 9.2\%. Similarly, Qwen3-VL (30B-A3B) prompted with our optimized prompt achieves 21.6 mAP, improving by 9.7\% over the baseline prompt (Figure \ref{fig:detpo_plot}). This suggests that effectively prompted generalist models can surpass purpose-built zero-shot specialists.

\textbf{DetPO Improves over Prior Prompt-Optimization Frameworks.} Table \ref{tab:rf20-vl} also shows that DetPO outperforms existing general-purpose prompt optimization techniques like GEPA \cite{agrawal2025gepa} and MIPROv2 \cite{opsahl2024optimizing} for object detection. When applied to Qwen3-VL (30B-A3B), GEPA provides a marginal improvement over the baseline, achieving 13.0 mAP, while MIPROv2 actually degrades to 10.7 mAP. In contrast, DetPO substantially improves the model's performance to 19.4 AP. This trend also holds for Gemini 3 Pro, where DetPO (26.4 mAP) outperforms GEPA (25.6 mAP) and MIPROv2 (25.0 mAP). These results indicate that our task-specific optimization approach is better suited for extracting robust object detection capabilities from generalist MLLMs than prior approaches.

\textbf{DetPO Transfers Across Different MLLMs.} We evaluate the generalizability of our prompt optimization approach across a diverse set of generalist detectors, including the Qwen2.5-VL and Qwen3-VL families. As shown in Table \ref{tab:transfer}, DetPO consistently improves detection performance over baseline prompts, regardless of model scale or architecture. For example, smaller models like Qwen2.5-VL (7B) improve from 6.2 to 9.1, while larger open models like Qwen2.5-VL (72B) improve from 10.4 to 15.7 mAP. Furthermore, VQA Score yields modest but consistent improvements across Qwen model variants (e.g., boosting Qwen3-VL 8B from 15.3 to 17.5 mAP), suggesting that precise confidence estimation is critical for maximizing detection performance in generalist models.

% \begin{figure}[t]
%     \centering
%     \includegraphics[width=0.5\linewidth]{figures/instruction_refinement_category_improvement.png}
%     \caption{\textbf{Iterative Accuracy Improvement.}}
%     \label{fig:detpo_plot}
% \end{figure}

{
\setlength{\tabcolsep}{1.1em}
\begin{table}[t]
\centering
\caption{\textbf{Ablation on Confidence Score}. We evaluate different approaches for estimating confidence scores with Qwen3-VL (30B-A3B). By default, we use the model's self-reported score (row 2). Notably, we find that using SigLIPv2's cosine similarity score between image crops of predicted boxes and class names for confidence scoring degrades overall performance (dropping from 19.4 to 16.4 mAP). In contrast, VQA Score effectively calibrates detection confidence, improving performance to 21.6 mAP.
\vspace{-1em}
} 
\label{tab:ablation_confidence}
\scalebox{0.70}{
\begin{tabular}{|lcccccccc|}
\hline

\rowcolor{gray!10}
\multicolumn{1}{|l|}{\textbf{Method}}                     & \multicolumn{1}{c|}{\textbf{A}} & \multicolumn{1}{c|}{\textbf{D}} & \multicolumn{1}{c|}{\textbf{F \& F}} & \multicolumn{1}{c|}{\textbf{I}} & \multicolumn{1}{c|}{\textbf{M}} & \multicolumn{1}{c|}{\textbf{S}} & \multicolumn{1}{c|}{\textbf{O}} & \textbf{All} \\ \hline
\multicolumn{1}{|l|}{Qwen3-VL \cite{Qwen3-VL} (30B-A3B) (C + I)}      & \multicolumn{1}{c|}{9.0}      & \multicolumn{1}{c|}{7.8}      & \multicolumn{1}{c|}{23.5}         & \multicolumn{1}{c|}{9.6}                    & \multicolumn{1}{c|}{0.7}                 & \multicolumn{1}{c|}{14.4}                & \multicolumn{1}{c|}{10.1}               &      11.9        \\ \hline
\multicolumn{1}{|l|}{\makecell[l]{\quad + Contrastive Prompt Optimization \\ \quad\quad (Self-Reported Score)}}      & \multicolumn{1}{c|}{13.8}      & \multicolumn{1}{c|}{18.6}      & \multicolumn{1}{c|}{34.6}         & \multicolumn{1}{c|}{19.7}                    & \multicolumn{1}{c|}{0.1}                 & \multicolumn{1}{c|}{21.8}                & \multicolumn{1}{c|}{16.4}               &     19.4         \\ \hline
\multicolumn{1}{|l|}{\quad \quad w/ SigLIPv2 \cite{tschannen2025siglip} Score}      & \multicolumn{1}{c|}{14.1}      & \multicolumn{1}{c|}{13.4}      & \multicolumn{1}{c|}{28.0}         & \multicolumn{1}{c|}{18.8}                    & \multicolumn{1}{c|}{0.0}                 & \multicolumn{1}{c|}{17.3}                & \multicolumn{1}{c|}{14.0}               &     16.4         \\ \hline
\multicolumn{1}{|l|}{\quad \quad w/ VQA Score \cite{lin2024evaluating}}       & \multicolumn{1}{c|}{16.1}      & \multicolumn{1}{c|}{25.2}      & \multicolumn{1}{c|}{36.5}         & \multicolumn{1}{c|}{20.1}                    & \multicolumn{1}{c|}{0.2}                 & \multicolumn{1}{c|}{25.7}                & \multicolumn{1}{c|}{18.4}               &     \textbf{21.6}         \\ \hline
\end{tabular}
}
\end{table}
}

\begin{figure}[t]
    \centering
    \includegraphics[width=0.95\linewidth]{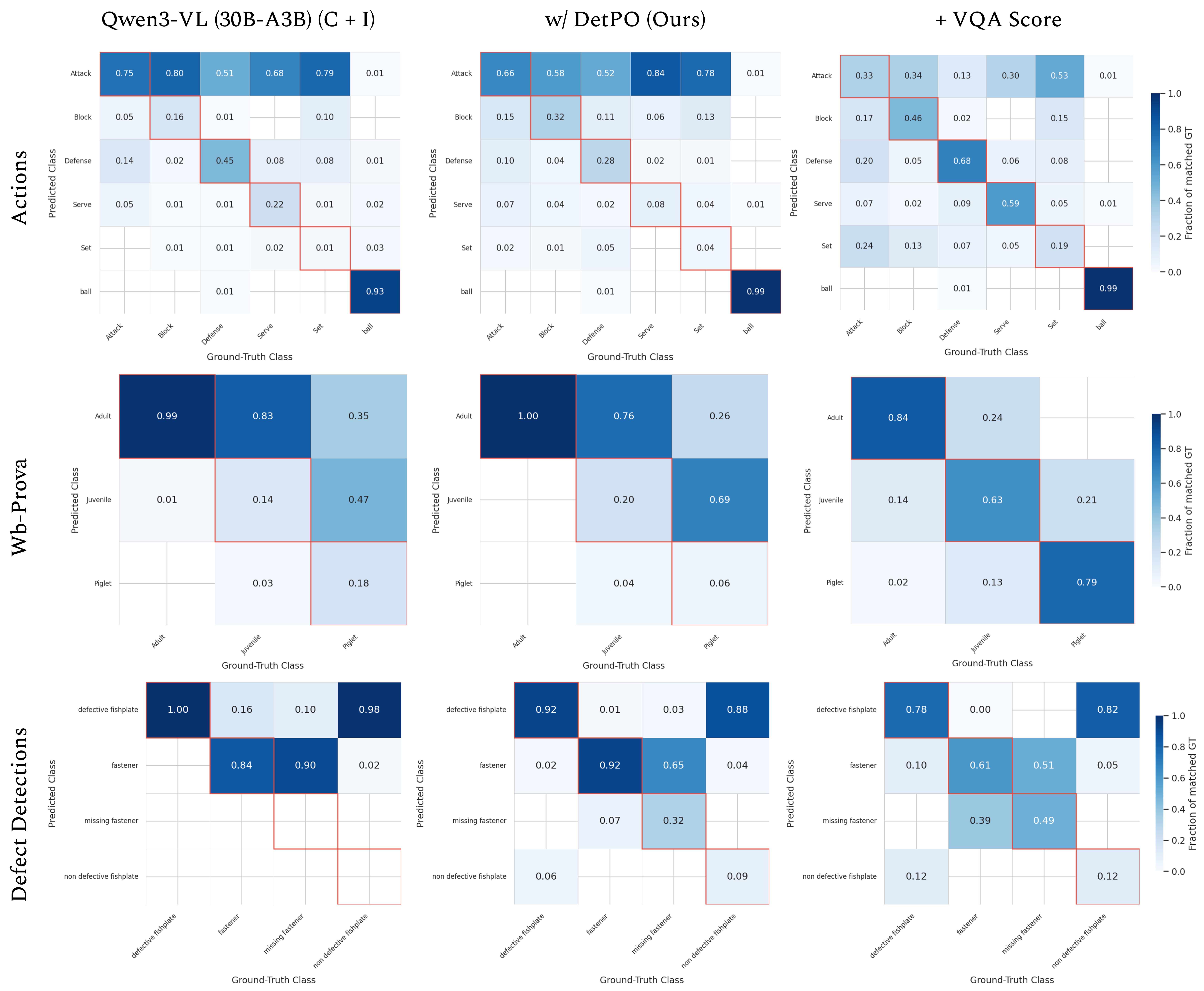}
    \caption{\textbf{Detection Confusion Matrix.} We compare  Qwen3-VL (30B-A3B), Qwen3-VL with DetPO, and with VQA Score across the Actions, Wb-Prova, and Defect Detections datasets. We find that DetPO and VQA Score consistently resolve baseline class imbalances. Notably, our proposed approach improves true positive rates for underrepresented classes ({\tt Juvenile}, {\tt Piglet}) and nuanced actions ({\tt Defense}, {\tt Serve}), while mitigating aggressive false positive predictions in defect detection.}
    \vspace{-2.5em}
    \label{fig:detpo_confusion}
\end{figure}

% \vspace{7em}
\textbf{Contrastive Prompt Optimization Reduces Class Confusion.}
We visualize a detection confusion matrix \cite{peri2023towards} for Qwen3-VL (30B-A3) in Figure \ref{fig:detpo_confusion}. Based on the confusion matrices, DetPO and VQA Score generally improve the model's ability to distinguish nuanced or underrepresented classes. The most dramatic improvement occurs in the Wb-Prova dataset, where the baseline almost entirely fails to identify {\tt Juvenile} and {\tt Piglet} boars, instead misclassifying them as {\tt Adult}. Further, VQA Score sigificantly improves their true positive rates on the diagonal to 63\% and 79\% respectively, effectively resolving the severe class imbalance. In the Actions dataset, the baseline struggles with specific actions like {\tt Block} (16\%), {\tt Defense} (45\%), and {\tt Serve} (22\%). Adding DetPO slightly improves {\tt Block} (32\%) and achieves near-perfect {\tt ball} recall (99\%), while VQA Score significantly helps distinguish nuanced actions, bringing {\tt Defense} to 68\% and {\tt Serve} to 59\%. Finally, in the Defect Detections dataset, the baseline suffers from aggressive overconfidence, misclassifying 98\% of {\tt non-defective fishplates} as defective. Incorporating DetPO and VQA Score helps mitigate false positives, incrementally improving the model's ability to correctly recognize non-defective parts (up to 12\%). We include examples of the original prompt, DetPO's initial prompt, an DetPO's optimized prompt for the {\tt soft plastic} class below. We highlight key details from each prompt in \textcolor{blue}{blue}.

% \greybox{\underline{Original Prompt:} `Serve' involves a player initiating play by striking the ball from behind the end line to send it over the net.}

% \greybox{\underline{DetPO Initial Prompt:} 'Serve': 'A player in the act of serving a volleyball, characterized by being near the back boundary line, with arms raised and contacting the ball in a service motion, typically from a standing or low-crouch position, distinct from airborne actions near the net such as spiking or setting, and not requiring full airborne elevation.}

% \greybox{\underline{DetPO Optimized Prompt:} 'Serve': 'A player in the act of serving a volleyball, characterized by being airborne or jumping near the back boundary line, with arms raised and contacting the ball in a controlled, overhand motion. The player is positioned behind the service line, and the action is initiated from the back of the court, distinct from airborne actions near the net such as spiking or setting, which involve a downward strike and occur closer to the net.}
\vspace{-1.5em}
\greybox{\underline{Original Prompt:} Soft plastic is often transparent or semi-transparent, featuring a flexible, wrinkled appearance, and have diverse visual appearances.}
\vspace{-2.75em}
\greybox{\underline{DetPO Initial Prompt:} `Soft plastic': Small, flexible, and thin sheets or bags made of translucent plastic material. Characterized by a smooth, \textcolor{blue}{shiny surface that reflects light}, often with crinkled or folded textures.}
\vspace{-2.75em}
\greybox{\underline{DetPO Optimized Prompt:} `Soft plastic': Thin, flexible, and translucent plastic material that appears in crumpled, or folded forms, often as sheets, bags, or loose fragments. Characterized by a smooth, shiny surface that reflects light, with crinkled or wrinkled textures. \textcolor{blue}{It may hold contents but lacks rigid structures, doesn't maintain fixed shape, and conforms to surfaces.}}
\vspace{-1em}

\textbf{Estimating Confidence Scores Improves Detection Accuracy.}
Table \ref{tab:ablation_confidence} evaluates different methods for confidence estimation applied to Qwen3-VL (30B-A3B). Our results suggest that self-reported confidence scores (row 2) provide a significant improvement over the baseline, which does not predict per-box confidences at all (where we assign each bounding box with a score of 1.0). We find that using SigLIPv2's cosine similarity score between image crops of predicted boxes and class names for confidence scoring degrades overall performance, dropping from 19.4 to 16.4 mAP, particularly hurting performance for Documents (18.6 to 13.4 mAP) and Flora \& Fauna (34.6 to 28.0 mAP). In contrast, we find that VQA Score proves to be highly effective; by leveraging the model's own logit scores, we can boost overall performance to 21.6 mAP. Furthermore, VQA Score demonstrates consistent improvements across all domains.

{
\setlength{\tabcolsep}{1.1em}
\begin{table*}[h]
\centering
\caption{\textbf{Black Box Prompting vs White-Box Fine-Tuning.} We compare black box prompting methods against fine-tuned open-weight models. While our DetPO framework significantly improves the spatial reasoning capabilities of frontier MLLMs like Gemini 3 Pro, full white-box fine-tuning of specialist models (e.g., GroundingDINO) currently performs the best. However, our analysis reveals a strong trendline as recent closed-source MMLMs like Gemini are far more performant. We suspect that DetPO applied to the next generation of frontier MMLMs will outperform all prior art.
} 
\label{tab:fine-tune}
\scalebox{0.70}{
\begin{tabular}{|lcccccccc|}
\hline
\rowcolor{gray!10}
\multicolumn{1}{|l|}{\textbf{Method}}                     & \multicolumn{1}{c|}{\textbf{A}} & \multicolumn{1}{c|}{\textbf{D}} & \multicolumn{1}{c|}{\textbf{F \& F}} & \multicolumn{1}{c|}{\textbf{I}} & \multicolumn{1}{c|}{\textbf{M}} & \multicolumn{1}{c|}{\textbf{S}} & \multicolumn{1}{c|}{\textbf{O}} & \textbf{All} \\ \hline
\rowcolor{gray!10}
\multicolumn{9}{|l|}{\textbf{Black Box Prompting}} \\ \hline 
\multicolumn{1}{|l|}{GroundingDINO \cite{liu2024grounding} (C)}                       & \multicolumn{1}{c|}{28.5}      & \multicolumn{1}{c|}{5.1}      & \multicolumn{1}{c|}{33.7}         & \multicolumn{1}{c|}{12.8}                    & \multicolumn{1}{c|}{0.4}                 & \multicolumn{1}{c|}{5.1}                & \multicolumn{1}{c|}{16.9}               &       16.8       \\ \hline
\multicolumn{1}{|l|}{Qwen3-VL \cite{Qwen3-VL} (8B)  (C + I)}      & \multicolumn{1}{c|}{7.1}      & \multicolumn{1}{c|}{7.3}      & \multicolumn{1}{c|}{24.8}         & \multicolumn{1}{c|}{9.3}                    & \multicolumn{1}{c|}{0.2}                 & \multicolumn{1}{c|}{10.2}                & \multicolumn{1}{c|}{10.9}               &      11.4        \\ \hline
%\multicolumn{1}{|l|}{\quad w/ DetPO (Ours)}      & \multicolumn{1}{c|}{8.3}      & \multicolumn{1}{c|}{19.1}      & \multicolumn{1}{c|}{30.3}         & \multicolumn{1}{c|}{13.7}                    & \multicolumn{1}{c|}{0.1}                 & \multicolumn{1}{c|}{14.2}                & \multicolumn{1}{c|}{12.2}               &       15.3       \\ \hline 
\multicolumn{1}{|l|}{\quad w/ DetPO (Ours) + VQA Score}      & \multicolumn{1}{c|}{12.3}      & \multicolumn{1}{c|}{24.2}      & \multicolumn{1}{c|}{32.3}         & \multicolumn{1}{c|}{13.5}                    & \multicolumn{1}{c|}{0.2}                 & \multicolumn{1}{c|}{17.2}                & \multicolumn{1}{c|}{14.3}               &    17.5         \\ \hline 
\multicolumn{1}{|l|}{Qwen3-VL \cite{Qwen3-VL} (30B-A3B)  (C + I)}      & \multicolumn{1}{c|}{7.1}      & \multicolumn{1}{c|}{7.3}      & \multicolumn{1}{c|}{24.8}         & \multicolumn{1}{c|}{9.3}                    & \multicolumn{1}{c|}{0.2}                 & \multicolumn{1}{c|}{10.2}                & \multicolumn{1}{c|}{10.9}               &      11.4        \\ \hline
%\multicolumn{1}{|l|}{\quad w/ DetPO (Ours)}     & \multicolumn{1}{c|}{13.8}      & \multicolumn{1}{c|}{18.6}      & \multicolumn{1}{c|}{34.6}         & \multicolumn{1}{c|}{19.7}                    & \multicolumn{1}{c|}{0.1}                 & \multicolumn{1}{c|}{21.8}                & \multicolumn{1}{c|}{16.4}               &     19.4         \\ \hline
\multicolumn{1}{|l|}{\quad w/ DetPO (Ours) + VQA Score}      & \multicolumn{1}{c|}{16.1}      & \multicolumn{1}{c|}{25.2}      & \multicolumn{1}{c|}{36.5}         & \multicolumn{1}{c|}{20.1}                    & \multicolumn{1}{c|}{0.2}                 & \multicolumn{1}{c|}{25.7}                & \multicolumn{1}{c|}{18.4}               &     21.6         \\ \hline
\multicolumn{1}{|l|}{Gemini 3 Pro \cite{DeepMind2025Gemini3Pro} (C + I + V)}      & \multicolumn{1}{c|}{27.0}      & \multicolumn{1}{c|}{26.7}      & \multicolumn{1}{c|}{31.3}         & \multicolumn{1}{c|}{26.2}                    & \multicolumn{1}{c|}{2.6}                 & \multicolumn{1}{c|}{26.9}                & \multicolumn{1}{c|}{13.3}               &       23.8       \\ \hline
%\multicolumn{1}{|l|}{\quad w/ DetPO$^*$ (Ours)}      & \multicolumn{1}{c|}{27.0}      & \multicolumn{1}{c|}{35.4}      & \multicolumn{1}{c|}{35.7}         & \multicolumn{1}{c|}{23.0}                    & \multicolumn{1}{c|}{3.8}                 & \multicolumn{1}{c|}{28.5}                & \multicolumn{1}{c|}{20.8}               &       26.4        \\ \hline
\multicolumn{1}{|l|}{\quad w/ DetPO (Ours) + VQA Score}      & \multicolumn{1}{c|}{26.2}      & \multicolumn{1}{c|}{35.7}      & \multicolumn{1}{c|}{35.4}         & \multicolumn{1}{c|}{23.3}                    & \multicolumn{1}{c|}{3.9}                 & \multicolumn{1}{c|}{28.2}                & \multicolumn{1}{c|}{20.4}               &       \textbf{26.3}      \\ \hline
\rowcolor{gray!10}
\multicolumn{9}{|l|}{\textbf{White-Box Fine-Tuning}} \\ \hline 
\multicolumn{1}{|l|}{Grounding-DINO (Fine-Tuned) \cite{robicheaux2025roboflow100}}                       & \multicolumn{1}{c|}{39.9}      & \multicolumn{1}{c|}{34.5}      & \multicolumn{1}{c|}{45.7}         & \multicolumn{1}{c|}{37.8}                    & \multicolumn{1}{c|}{23.3}                 & \multicolumn{1}{c|}{26.3}                & \multicolumn{1}{c|}{24.7}               &      \textbf{33.4}        \\ \hline
\multicolumn{1}{|l|}{Qwen3-VL \cite{Qwen3-VL} (8B)  (C + I) (LoRA)}      & \multicolumn{1}{c|}{7.9}      & \multicolumn{1}{c|}{13.7}      & \multicolumn{1}{c|}{26.2}         & \multicolumn{1}{c|}{10.2}                    & \multicolumn{1}{c|}{0.3}                 & \multicolumn{1}{c|}{11.0}                & \multicolumn{1}{c|}{12.3}               &      13.2        \\ \hline
\end{tabular}
}
\end{table*}
}

\textbf{In-Context Prompting Under Performs Fine-Tuning.}
Table~\ref{tab:fine-tune} shows that black-box prompt optimization still underperforms ``white-box'' fine-tuning of open-weight models. We note that this is surprising given that for other recognition tasks like visual question answering (VQA), frontier closed-source models resoundingly outperform their open-weight counterparts~\cite{naturalbench}. Currently, this is not the case for spatial tasks such as object detection. However, our analysis reveals a strong trendline as recent closed-source MMLMs like Gemini are far more performant. We suspect that DetPO applied to the next generation of frontier MMLMs will outperform fine-tuned open-weight models such as GroundingDINO.

\begin{figure}[t]
    \centering
    \includegraphics[width=\linewidth]{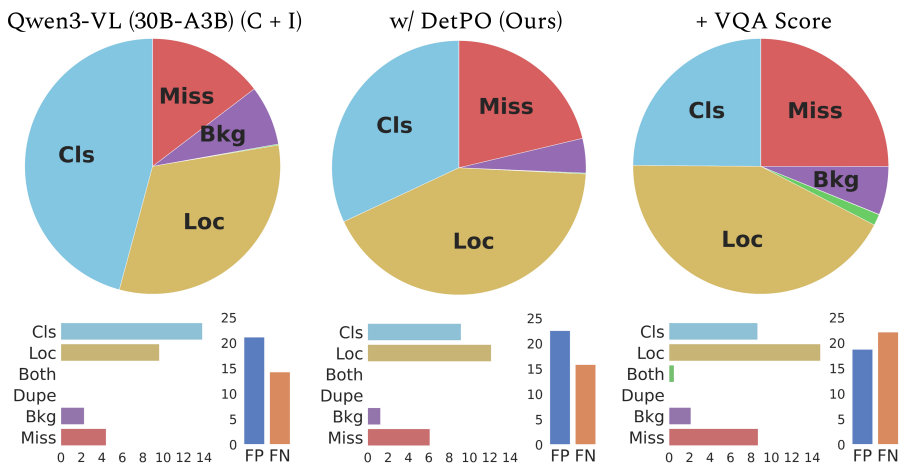}
    \caption{\textbf{Detection Errors.} We diagnose errors in the baseline Qwen3-VL (30B-A3B) model (\textbf{left}), the proposed DetPO method (\textbf{center}), and DetPO + VQA Score (\textbf{right}) with TIDE \cite{bolya2020tide}. The top row shows the relative distribution of error types, while the bottom row describes the absolute error counts and the overall false positive (FP) versus false negative (FN) rates. DetPO notably reduces classification errors compared to the baseline. While adding in VQA Score successfully reduces overall false positives, it shifts the primary error bottleneck to localization (Loc) and significantly increases missed detections (FN).}
    \label{fig:detpo_erros}
\end{figure}

\textbf{Analysis of DetPO Errors.}
We systematically analyze model errors using TIDE \cite{bolya2020tide} in Figure \ref{fig:detpo_erros}. Our analysis shows that DetPO directly addresses the baseline model's primary weakness by substantially reducing the class confusion rate. We attribute this improvement to contrastive prompt refinement, which explicitly highlights discriminative features between classes. However, re-scoring detections with VQA Score creates a new trade-off: while classification errors remain low and false positives decrease overall, localization errors and missed detections increase substantially. We posit that this is because some true positive detections are incorrectly down-weighted. Overall, DetPO effectively reduces both false positives and mis-classifications, significantly improving detection accuracy.

\begin{figure}[t]
    \centering
    \includegraphics[width=\linewidth]{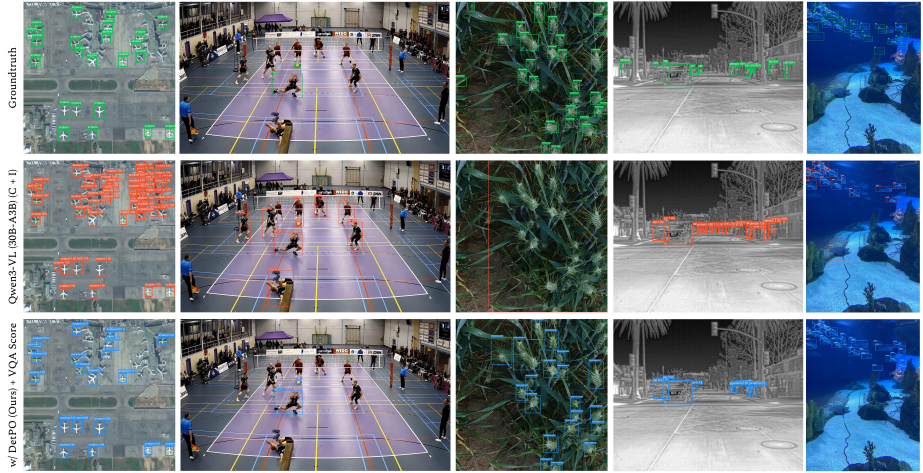}
    \caption{\textbf{Qualitative Results.} The baseline Qwen3-VL model suffers from high false positive rates (dense, overlapping boxes) and poor recall in complex environments. In contrast, our proposed method mitigates these issues, significantly reducing erroneous predictions while successfully recovering missed objects (like {\tt wheat heads} and {\tt fish}). Best viewed zoomed in.}
    \label{fig:qualitative}
\end{figure}

% \vspace{5em}
\textbf{Qualitative Examples.}
We visualize qualitative examples in Figure \ref{fig:qualitative}. Our approach significantly outperforms the baseline Qwen3-VL model across a diverse set of challenging domains, including aerial imagery, sports, agriculture, thermal imaging, and underwater scenes. The baseline model generates many dense, overlapping, and erroneous bounding boxes in both the aerial airplane scene and the thermal street view, leading to many false positives. Further, it suffers from poor recall, failing almost entirely to detect the {\tt wheat heads}. In contrast, the proposed method effectively mitigates these shortcomings, suppressing false positives to produce precise, well localized predictions that closely align with the ground truth. It also successfully recovers objects missed by the baseline (such as the {\tt wheat heads} and smaller {\tt fish}), demonstrating strong performance across diverse environments.

\textbf{Limitations and Future Work}
While Detection Prompt Optimization (DetPO) effectively bridges the gap between zero-shot generalization and task-specific adaptation, the iterative prompt refinement process introduces computational overhead similar to training a specialist model. However, it is important to note that this optimization cost is only incurred once during the initial prompt discovery phase; subsequent inference calls using the optimized prompt are no more expensive than standard MLLM inference calls. We expect that future innovations in speeding up inference (e.g. vLLM \cite{kwon2023efficient}) will continue to reduce wall-clock time. Further, we find that DetPO fails to improve performance on medical datasets, likely because weak base-model performance limits the benefits of prompt optimization. More generally, DetPO appears most effective in the intermediate regime where models are capable but not yet saturated. Furthermore, our experiments demonstrating the effectiveness of VQA Score suggest that MLLM self-reported confidence scores are suboptimal. Future work may improve performance further by investigating better confidence calibration mechanisms. Finally, evaluating closed-source MLLMs like Gemini 3 Pro via APIs introduces potential training data leakage concerns. By evaluating these models on specific training examples, there is a risk of images and optimized prompts being used in future pre-training mixtures.

\section{Conclusion}
In this paper, we demonstrate that current MLLMs struggle to effectively leverage few-shot multi-modal in-context examples for object detection. To address this limitation, we present Detection Prompt Optimization (DetPO) as a simple yet effective solution. By iteratively refining text-only prompts and incorporating visual feedback from both positive and negative detections, DetPO bridges the gap between zero-shot generalization and task-specific adaptation without gradient-based training. Our experiments on RF20-VL and LVIS demonstrate that DetPO improves concept alignment, reduces false positive detections through improved confidence calibration, and yields consistent performance improves across diverse domains and generalist MLLMs. Our results suggest that prompt-level optimization can serve as a practical alternative to fine-tuning in cases where it is infeasible (e.g., closed-source APIs) or prohibitively expensive (e.g., state-of-the-art open-weights MLLMs), enabling generalist MLLMs to better adapt to novel classes, modalities, and tasks encountered in real-world scenarios.

%Our analysis also shows the surprising result that closed-source frontier models currently underperform open-weight models when fine-tuned. We posit that this is due to highly spatial nature of the object detection task. We believe that this discrepancy will be addressed in future generations of frontier MMLMs, suggesting black-box optimizers will still be useful.

\section*{Acknowledgments}
This work was supported in part by the NSF GRFP (Grant No. DGE2140739).

\newpage
\clearpage

\bibliographystyle{splncs04}
\bibliography{main}

\newpage 

\appendix 

\section{Baseline Implementation Details}
\label{sec:implementation_details}
We present additional implementation details to reproduce our baseline experiments below.

\textbf{GroundingDINO} is a text-promptable vision-language model designed for open-set object detection. It combines the DINO transformer-based object detector with language grounding so that textual queries like ``a brown dog'' or ``traffic light'' guide the detection process. The model aligns image regions with text embeddings to produce bounding boxes for objects that match the query, even if those categories were not explicitly included in training. We use GroundingDINO \cite{liu2023grounding} with pretrained weights from mmdetection (MM-GroundingDINO-L*). We prompt the model with all the class names combined into a single prompt. For ``white-box'' experiments, we fine-tune GroundingDINO on each few-shot dataset for 1000 iterations with a batch size of 4 and a learning rate of 3e-4. We resize all images to (640, 1333) and don't use any additional data augmentations.

\textbf{MQ-GLIP} proposes a learnable module that enables multi-modal prompting. We choose GLIP with a SWIN-L backbone as the underlying detection model for our experiments. We use the model checkpoint trained on Objects365, FourODs, GoldG, and Cap24M. Lastly, we use class names as the text prompts and few-shot visual examples as visual prompts.

\textbf{YOLO-E} presents a unified open-vocabulary framework that integrates re-parameterizable region-text alignment (RepRTA) and semantic-activated visual prompt encoding (SAVPE). We select the YOLOv8-based architecture as the underlying detection model. We utilize the model checkpoint pre-trained on large-scale grounding datasets including Objects365 and GoldG. Lastly, we use class names as text prompts and few-shot training images as visual prompts to enable zero-shot detection.

\textbf{SAM3} is a recent open-vocabulary image and video segmentation model. Unlike traditional segmentation models that require predefined object classes, SAM3 can detect and generate pixel-level masks for any object described by a concept, using prompts such as natural-language text (e.g., ``red car''), example images, or interactive clicks. We prompt for each class independently since SAM3 does not natively support multi-class detection.

\textbf{LLMDet} integrates LLM supervision into GroundingDINO's architecture to improve open-vocabulary object detection. Unlike traditional detectors that can only recognize a fixed set of predefined classes, LLMDet leverages the semantic knowledge of LLMs to learn from both region-level descriptions and image-level captions, enabling it to detect objects described in natural language, even if those object categories were not present during training. This language-guided supervision helps the model generalize better to new or rare object categories and perform zero-shot detection.

\section{Prompt Optimization Baseline Details}

We use DSPy \cite{khattab2023dspy} to define a object detection program that takes an image and text prompt as input and produces a JSON string of bounding box detections. The detection prompt is automatically constructed from each dataset's categories and README metadata (i.e., per-class descriptions and annotator instructions), instructing the model to detect all target classes simultaneously. For Gemini 3 Pro, bounding boxes use the native \texttt{[ymin, xmin, ymax, xmax]} format normalized to $[0, 1000]$; for Qwen 3-VL, we use \texttt{[x1, y1, x2, y2]} in the same coordinate space. We optimize per-image F1 at IoU $\geq 0.5$ with greedy matching. The metric provides structured text feedback (e.g. precision, recall, and specific error descriptions) to guide prompt evolution.

\textbf{GEPA} automatically improve prompts or instructions by estimating gradient-like signals from model outputs and feedback to iteratively update prompts. The algorithm evaluates how changes to a prompt affects task performance and then adjusts the prompt in the direction that improves results. This automatically refines prompts for better accuracy, reasoning quality, or task performance while treating the MLLM as a black box. We run GEPA \cite{agrawal2025gepa} with the \texttt{light} budget preset, generating 6 candidate prompts over ${\sim}10$ evolutionary trials. GEPA reflects on minibatches of 3 training examples per step using the same base model as the reflection LM (temperature 0.8), and selects the candidate with the highest validation F1. 

\textbf{MiPROv2} is an automated prompt optimization algorithm that improves LLM performance by generating, evaluating, and combining multiple candidate instructions and demonstrations. It searches over different prompt structures such as task instructions, examples, and formatting and selects the best configuration based on validation performance. We run MIPROv2 \cite{opsahl2024optimizing} with the same \texttt{light} budget preset as GEPA. Unlike GEPA, MIPROv2 proposes instruction candidates via LLM-based generation rather than evolutionary reflection, and selects among them using Bayesian optimization over validation scores.

\section{DetPO Implementation Details}
\label{sec:algo}
We present additional implementation details to reproduce our DetPO experiments below and open source our code on \href{https://github.com/ggare-cmu/DetPO}{GitHub}.

\textbf{DetPO (Detection Prompt Optimization)} is a single-class iterative prompt optimization framework that automatically refines per-class natural-language descriptions for MLLM-based object detection. DetPO uses the MLLM as both a detector and a prompt critic, iteratively improving textual class definitions by reasoning over its own detection errors. Given a set of object classes $\mathcal{C}$, a training split, and a maximum number of iterations $T_{max}$, DetPO produces a refined natural-language class definition $Pc$ for each class $c$ that maximizes detection performance on a held-out validation set. The full procedure is summarized in Algorithm~\ref{alg:mmpo} and is described in detail below.
 
\textbf{Stage 1: Initial Prompt Generation.}
\label{sec:stage1}
The goal of stage~1 is to bootstrap a class definition from labeled visual examples before running inference on unannotated test images. It consists of two parts:
 
\textsc{SummarizePositive}. All ground truth instances in the training set containing class $c$ are annotated with green bounding boxes. Next, the MLLM is prompted with all annotated images and is asked to identify the consistent visual characteristics of the highlighted objects. This produces a concise class definition. A seed description from the dataset README provides a prior that grounds the initial generation:
\begin{align}
  \Pc \;\leftarrow\; \mathrm{MLLM}\!\left(\,
    \bigl\{\,\mathrm{DrawGreen}(x,\,b_c)\;\big|\; x \in \mathcal{X}^+(c)\,\bigr\}
  \,\right).
  \label{eq:summarize}
\end{align}
 
\textsc{RefineContrastive}. For every other class $c^- \neq c$, one example image containing $c^-$ is randomly sampled. That image is annotated with a red bounding box around the $c^-$ instance. We prompt the MLLM with both a negative example from $c^-$ and positive example from class $c$. The MLLM then identifies the key features distinguishing the object inside the green box from the red box, and produces an updated $\Pc$ that excludes $c^-$ while including $c$:
\begin{align}
  \Pc \;\leftarrow\; \mathrm{RefineContrastive}\!\left(\Pc,\;
    \mathrm{DrawGreen}(x^+_c),\;
    \mathrm{DrawRed}(x^-_{c^-})
  \right).
  \label{eq:contrastive}
\end{align}
This loop iterates over all $|\mathcal{C}|-1$ negative classes, progressively
sharpening $\Pc$ to exclude visually similar but semantically distinct objects.
 
% ─────────────────────────────────────────────────────────────────────────────
\textbf{Stage 2: Iterative Error-Driven Refinement.}
\label{sec:stage2}
Stage~2 runs a closed-loop optimization in which $\Pc$ is evaluated on the training split. We identify the worst false positive (FP) and worst false negative (FN) detections and update $\Pc$ to address each error type.
 
\textsc{IdentifyErrors}.
At each iteration $t$, we run inference on the training split using $\Pc$ and compute COCO-style metrics. Per-detection error scores are derived as follows:
 
\textit{False Positive Score.}
For each predicted box that does not overlap with any ground truth box of class $c$, the false positive error is proportional to the model's confidence and its overlap with the ground truth boxes of other classes:
\begin{equation}
  \varepsilon_{\mathrm{FP}}(d) \;=\; s_d \cdot \max\!\bigl(0.2,\;
    \mathrm{IoU}(b_d, b^*_{\neq c})\bigr),
  \label{eq:fp_score}
\end{equation}
where $s_d$ is the detection confidence and $b^*_{\neq c}$ is the nearest ground truth box of any class other than $c$.
 
\textit{False Negative Score.}
For each ground truth box of class $c$ that is not matched with a prediction, the false negative error reflects how far the best matching prediction falls from a correct detection:
\begin{align}
  \sigma(g) &\;=\; \max_{d}\;\bigl[s_d \cdot \mathrm{IoU}(b_d, g)\bigr],
  \label{eq:best_score}\\
  \varepsilon_{\mathrm{FN}}(g) &\;=\; 1 - \sigma(g).
  \label{eq:fn_score}
\end{align}
 
The worst false positive is selected as $\arg\max_d \,\varepsilon_{\mathrm{FP}}(d)$ and the
worst false negative as $\arg\max_g \,\varepsilon_{\mathrm{FN}}(g)$. To encourage diversity, previously selected images are excluded from the candidate pool. We use the best match (e.g. highest-scoring correct detection), worse false positive (e.g. highest error false positive), and worse false negative (e.g. highest error false negative) as visual evidence.

\textsc{RefineInclude}.
The MLLM then receives the best match (\greenbox{}) and worst false negative (\bluebox{}) images alongside the current $\Pc$. The MLLM identifies features shared by both objects and produces a definition that would also detect the blue instance. False negative refinement is applied before false positive refinement within each iteration so that the broadened definition is subsequently tightened:

\begin{align}
  \Pc \;\leftarrow\; \mathrm{RefineInclude}\!\left(\Pc,\;
    x^+_{\greenbox{}},\; x^{\mathrm{FN}}_{\bluebox{}}
  \right).
  \label{eq:refine_fn}
\end{align}

\textsc{RefineExclude}.
The MLLM receives the best match (\greenbox{}) and worst false positive (\redbox{}) images alongside the updated $\Pc$. The MLLM is asked to identify features that distinguish the two objects and produce a definition that would reject the worse false positive:
\begin{align}
  \Pc \;\leftarrow\; \mathrm{RefineExclude}\!\left(\Pc,\;
    x^+_{\greenbox{}},\; x^{\mathrm{FP}}_{\redbox{}}
  \right).
  \label{eq:refine_fp}
\end{align}

\textit{Conservative Update Rule and Early Stopping}
After each iteration, we compare the mAP of the updated prompt to the previous iteration's training mAP. If performance decreases, $\Pc$ is reverted to the previous best prompt
(\texttt{prev\_instructions}). The best global prompt is tracked separately:
\begin{align}
  \Pc^{(t)} &\;\leftarrow\;
  \begin{cases}
    \Pc^{(t)} & \text{if } \mathrm{mAP}(\Pc^{(t)}) \geq \mathrm{mAP}(\Pc^{(t-1)})\\
    \Pc^{(t-1)} & \text{otherwise (revert)}
  \end{cases}
  \label{eq:revert}\\[4pt]
  \Pc^* &\;\leftarrow\; \arg\max_{t}\; \mathrm{mAP}(\Pc^{(t)},\;\mathcal{D}_{\mathrm{train}}).
  \label{eq:best}
\end{align}

The loop terminates at $t = \Tmax$ or when all sub-sample evaluation scores are already perfect (early stopping). 
 
\textbf{Stage 3: Validation Set Candidate Selection}
\label{sec:stage3}
After $\Tmax$ iterations, Stage~3 selects the final prompt from a set of
candidates by evaluating each on the held-out validation split.
 
\textsc{GenerateAlternative.}
We also ask the MLLM to generate an additional candidate prompt $\Pc^{\mathrm{alt}}$ to refine the best prompt without any visual examples. This allows the MLLM to remove dataset-specific artifacts:
\begin{align}
  \Pc^{\mathrm{alt}} \;\leftarrow\; \mathrm{MLLM}\!\left(\Pc^*,\;
    \textit{``Refine for generalization''}\right).
  \label{eq:alt}
\end{align}
 
\textbf{Candidate Evaluation and Selection.}
We evaluate five candidate prompts on $\mathcal{D}_{\mathrm{val}}$ using COCO
mAP. The highest-scoring candidate is selected as the final output for class $c$:
\begin{align}
  \Pc^{\mathrm{final}} \;\leftarrow\; \arg\max_{p \,\in\, \mathcal{P}}
    \;\mathrm{mAP}(p,\;\mathcal{D}_{\mathrm{val}}),
  \label{eq:val_select}
\end{align}
where $\mathcal{P} = \{\Pc^{(0)},\; \Pc^{(1)},\; \Pc^*,\; \Pc^{(\Tmax)},\;
\Pc^{\mathrm{alt}}\}$. $\Pc^{(0)}$ is the dataset provided seed prompt, $\Pc^{(1)}$ is the Stage~1
prompt, $\Pc^*$ is the best training iteration prompt, $\Pc^{(\Tmax)}$ is the final training iteration prompt, and $\Pc^{\mathrm{alt}}\}$ is the
alternative generated prompt.
 
\textbf{Model Specific Details.} We conduct all Qwen2.5-VL experiments using the ``qwen2.5-vl-72b-instruct'' model and ``qwen2.5-vl-7b-instruct'' model. Similarly, we conduct all Qwen3-VL experiments using the ``qwen3-vl-8b-instruct'' and ``qwen3-vl-30b-a3b-instruct'' models. We prompt the model based on guidelines from Qwen's official documentation. For Gemini 3 Pro experiments, we generate initial DetPO prompts using \textit{gemini-3-flash-preview}. Since the Gemini API does not expose token-level log probabilities, we perform score calibration with VQAScore using Qwen3-VL~\cite{Qwen3-VL} (30B-A3B).

\RestyleAlgo{ruled}
\SetKwComment{Comment}{/* }{ */}
\begin{algorithm}[H]
\caption{DetPO Algorithm}
\label{alg:mmpo}
\begin{pythonic}
# Single-Class Iterative Prompt Tuning
def tune_prompt(classes, train_set, T_max):
    refined_prompt = ""

    # Stage 1: Initial prompt
    P_c = SummarizePositive(c)
    for c_neg in classes:
        if c_neg == c: continue
        P_c = RefineContrastive(P_c, sample_positive(c), sample_negative(c_neg))

    # Stage 2: Iterative refinement
    for t in range(T_max):
        FP, FN = IdentifyErrors(ModelDetect(train_set, P_c))

        # Select worst errors
        FP_err = argmax(FP, key="confidence")
        FN_err = argmin(FN, key="IoU")

        # Update prompt
        x_pos = sample_positive(c)
        P_c = RefineExclude(P_c, x_pos, FP_err)
        P_c = RefineInclude(P_c, x_pos, FN_err)

        refined_prompt = P_c

        # Early stop
        if has_converged(EvaluatePrompt(P_c, train_set)):
            break

    # Stage 3: Val set based selection
    P_c_alt = GenerateAlternative(refined_prompt)
    candidates = [P_c_initial, refined_prompt, P_c_alt]
    refined_prompt = argmax(candidates, key=lambda p: EvaluatePrompt(p, val_set))
    
    return refined_prompt
    
\end{pythonic}
\end{algorithm}

\newpage
\section{Prompts}
\label{sec:prompts}
We improve Qwen3-VL's base prompt through small-scale validation on multiple datasets and select the best prompt:

% %Baseline Prompts - shruti
% Just class names - f"Locate all of the following objects: {category_prompt} in the image and output the coordinates in JSON format like {{\"bbox_2d\":[x1,y1,x2,y2],\"label\":\"class_name\"}}."
% Just instructions - f"Locate all of the following objects: {category_prompt} in the image and output the coordinates in JSON format like {{\"bbox_2d\":[x1,y1,x2,y2],\"label\":\"class_name\"}}.\n\nUse the following annotator instructions to improve detection accuracy:\n{instructions}\n"
% Instructions + images - f"Locate all of the following objects: {category_prompt} (each of those is a separate class) in the image and output the coordinates in JSON format like {{\"bbox_2d\":[x1,y1,x2,y2],\"label\":\"class_name\"}}."

\textbf{System Prompt}
% \greybox{``You are a helpful assistant capable of object detection.''}
\greybox{
\ttfamily\small
``You are a helpful assistant capable of object detection.''
}

\textbf{Multi-Class Detection Prompt}
% \greybox{``Locate all of the following objects: {category_prompt} in the image and output the coordinates in JSON format like {{\"bbox_2d\":[x1,y1,x2,y2],\"label\":\"class_name\"}}.''}
% \begin{tcolorbox}[width=\columnwidth]
% \texttt{Locate all of the following objects: {category\_prompt} in the image and output the coordinates in JSON format like {{\"bbox\_2d\":[x1,y1,x2,y2],\"label\":\"class\_name\"}}.}
% \end{tcolorbox}
\greybox{
\ttfamily\small
``Locate all of the following objects: \{\texttt{category\_prompt}\} in the image and output the coordinates in JSON format like \{\{"bbox\_2d":[x1,y1,x2,y2],
"label":"class\_name"\}\}.``
}

\textbf{Single-Class Detection Prompt}
% \greybox{``Locate every \{class name\} in the image and output the coordinates in JSON format.''}
\greybox{
\ttfamily\small
``Locate every \{class name\} in the image and output the coordinates in JSON format.''
}
% \textbf{Gemini 2.5 Pro.} We conduct all experiments using the Gemini API with the ``gemini-2.5-pro-preview-03-25'' model. We prompt the model based on guidelines from Gemini's official documentation,  but also improve the base prompt through small-scale validation on multiple datasets and select the best prompt:

% \textbf{System Prompt}\greybox{``Return bounding boxes as a JSON array with labels. Never return masks or code fencing.''}

% \textbf{Multi-Class Detection Prompt}\greybox{``Detect the 2d bounding boxes of the following objects: \{class names\}''}

% \textbf{Single-Class Detection Prompt}
% \greybox{``Detect all 2d bounding boxes of \{class name\}.''}

\textbf{Prompting with Rich Textual Instructions}
% To evaluate Qwen with dataset-specific annotator instructions, we appended the following prompt after our main prompt:

% \greybox{``Locate all of the following objects: {category_prompt} in the image and output the coordinates in JSON format like {{\"bbox_2d\":[x1,y1,x2,y2],\"label\":\"class_name\"}}.\n\nUse the following annotator instructions to improve detection accuracy:\n{instructions}''}
% \begin{tcolorbox}[width=\columnwidth]
% \texttt{Locate all of the following objects: {category\_prompt} in the image and output the coordinates in JSON format like \{\"bbox\_2d\":[x1,y1,x2,y2],\"label\":\"class\_name\"\}.\textbackslash n\textbackslash nUse the following annotator instructions to improve detection accuracy:\textbackslash n\{instructions\}}
% \end{tcolorbox}
\greybox{
\ttfamily\small
``Locate all of the following objects: \{\texttt{category\_prompt}\} in the image and output the coordinates in JSON format like \{\{"bbox\_2d":[x1,y1,x2,y2],
"label":"class\_name"\}\}.

Use the following annotator instructions to improve detection accuracy:
\{\texttt{instructions}\}'' 
}

We include the rich textual description for all classes when using the multi-class detection prompt. In contrast, we only append the relevant class description when using the single-class detection prompt.

\newpage

\textbf{Prompting with Few-Shot Visual Examples}
% \begin{tcolorbox}[width=\columnwidth]
% Locate all of the following objects: \texttt{\{category\_prompt\}} (each of those is a separate class) in the image and output the coordinates in JSON format like \texttt{\{\{\"bbox\_2d\":[x1,y1,x2,y2],\"label\":\"class\_name\"\}\}}.
% \end{tcolorbox}
\greybox{
\ttfamily\small
``Locate all of the following objects: \{\texttt{category\_prompt}\} (each of those is a separate class) in the image and output the coordinates in JSON format like \{\{"bbox\_2d":[x1,y1,x2,y2],
"label":"class\_name"\}\}.``
}

\textbf{Initial Class Definition Generation Prompt}
% \begin{tcolorbox}[width=\columnwidth]
% Analyze the following images and describe the subjects or objects highlighted in green bounding boxes. 
%             Identify and summarize the key visual characteristics that are consistently observed across these objects. 
%             Emphasize the distinctive features that clearly differentiate this object class from other elements in the scene.

%             Your goal is to produce a concise, clear, detailed, and generalizable definition that enables accurate recognition of this object class in future images and makes it easily distinguishable from other objects. 
%             Do not mention bounding boxes, colors, or any annotation details in your response.
% \end{tcolorbox}
\greybox{
\ttfamily\small
Analyze the following images and describe the subjects or objects highlighted in green bounding boxes. 
Identify and summarize the key visual characteristics that are consistently observed across these objects. 
Emphasize the distinctive features that clearly differentiate this object class from other elements in the scene.

Your goal is to produce a concise, clear, detailed, and generalizable definition that enables accurate recognition of this object class in future images and makes it easily distinguishable from other objects. 
Do not mention bounding boxes, colors, or any annotation details in your response.
}

\newpage

\textbf{Class Definition Refinement using False-Positive Prompt}

\greybox{
\ttfamily\small
Analyze the image carefully and identify the key visual differences between the object shown in the green bounding box and the one shown in the red bounding box.

Follow the following steps:

Step-1. Describe the distinguishing visual characteristics that set apart the object in the green bounding box from the object in the red bounding box.

Step-2. Based on these distinguishing traits, formulate a clear and descriptive class definition for the object in the green bounding box. This definition should focus on its unique visual and contextual features that help differentiate it from the object in the red bounding box.

Step-3. Compare your new class definition with the existing definition of the `\texttt{\{class\_name\}}` class:

Current class definition of the `\texttt{\{class\_name\}}` class:

\texttt{\{current\_instructions\}}

Step-4. Synthesize both definitions to produce an improved, more precise descriptive class definition for the `\texttt{\{class\_name\}}' class. The updated definition should make it easier to accurately identify true instances of the `\texttt{\{class\_name\}}` class while reducing false positives similar to the one seen in the red bounding box.

Note: Do not mention bounding boxes, colors, or image annotations in your response. The updated class definition should be a textual description of the `\texttt{\{class\_name\}}` class objects.

Return the final updated class definition as descriptive text in the following format:

\texttt{```python \{\{`\{class\_name\}': <updated definition>\}\}```}
}

\newpage
\textbf{Class Definition Refinement using False-Negative Prompt}

\greybox{
\ttfamily\small
Analyze the image carefully and identify the key visual similarities between the object shown in the green bounding box and the one shown in the blue bounding box.

Follow the following steps:

Step-1. Describe the similar visual characteristics that set apart the object in the green bounding box from the object in the blue bounding box.
            
Step-2. Based on these similarity traits, formulate a clear and descriptive class definition for the object in the blue bounding box as well as the object in the green bounding box. This definition should focus on its unique visual and contextual features that help identify both instances of the object in the green and blue bounding boxes.
            
Step-3. Compare your new class definition with the existing definition of the '\texttt{\{class\_name\}}' class provided below:

Current class definition of the `\texttt{\{class\_name\}}' class:

\texttt{\{current\_instructions\}}

Step-4. Synthesize both definitions to produce an improved, more precise descriptive class definition for the '\texttt{\{class\_name\}}' class. The updated definition should make it easier to accurately identify all true instances of the '\texttt{\{class\_name\}}' class similar to the one seen in the blue and green bounding boxes.

Note: Do not mention bounding boxes, colors, or image annotations in your response. The updated class definition should be a textual description of the '\texttt{\{class\_name\}}' class objects.

Return the final updated class definition as descriptive text in the following format:

\texttt{```python \{\{`\{class\_name\}': <updated definition>\}\}```}
}

\newpage
\textbf{Generate Alternative Class Definition}
\greybox{
\ttfamily\small
Refine the class definition for the `\texttt{\{class\_name\}}' category.

Objective:
Produce a concise, precise, and generalizable definition that enables reliable recognition of `\texttt{\{class\_name\}}' instances across diverse images. The definition should clearly distinguish this class from visually or functionally similar object categories.

Current definition:
\texttt{\{best\_instructions\}}

Guidelines:
- Focus on intrinsic, stable characteristics such as structure, shape, components, function, and typical physical configuration.
- Ensure the description is detailed enough for accurate visual identification, yet broadly applicable across variations.
- Do NOT mention bounding boxes, colors, image annotations, or dataset-specific context.
- Avoid referencing specific images or examples.
- Output only a textual class definition.

Return the result strictly in the following format: 

\texttt{```python \{\{`\{class\_name\}': <updated definition>\}\}```}
}

\newpage
\textbf{DetPO Single Class Detection Prompt}
\greybox{
\ttfamily\small
Identify and localize all instances of `\texttt{\{class\_name\}}' in the image.

Output Requirements:
- Return valid JSON only. Do not include explanations or extra text. \\
- Output a ranked list of detections sorted by confidence (highest first). \\
- Include at most 20 detections. \\
- If no objects are detected, return an empty list \texttt{[]}.

For each detection, provide: \\
- \texttt{"bbox\_2d": [x1, y1, x2, y2]} \\
\quad * Pixel coordinates. \\
\quad * (x1, y1) = top-left corner. \\
\quad * (x2, y2) = bottom-right corner. \\
- \texttt{"label": "\texttt{\{class\_name\}}"} \\
- \texttt{"score":} float confidence score from \texttt{0.0} (lowest) to \texttt{1.0} (highest).

Additional Constraints: \\
- Only include detections that clearly correspond to \texttt{\{class\_name\}}. \\
- Avoid duplicate or highly overlapping boxes for the same object. \\
- Follow these annotator instructions to improve detection accuracy:

\texttt{\{dataset\_instructions\}}

Return a JSON list in the following format: \\

\texttt{[} \\
\texttt{\{} \\
\quad \texttt{"bbox\_2d": [x1, y1, x2, y2],} \\
\quad \texttt{"label": "\texttt{\{class\_name\}}",} \\
\quad \texttt{"score": 0.95} \\
\texttt{\}} \\
\texttt{]}
}

% \newpage
\textbf{Confidence Estimation using VQA Score Prompt}
\greybox{
\ttfamily\small
Given the `\texttt{\{prompt\}}' class defined as follows: \{\texttt{dataset\_instructions}\}

Is the main subject or object being referred to as `\texttt{\{prompt\}}' located inside the red bounding box in the image? Please answer Yes or No. Note: The object should be entirely inside the bounding box, with no part outside, and it must be the only object present inside - no other objects should appear within the box.
}

\newpage

\section{LVIS Rare 50 10-Shot Benchmark Results}
\label{sec:lvis}
To further evaluate few-shot detection performance, we constructed a subset of LVIS by sampling the 50 least frequent classes in the validation set that contain at least 10 examples each. While many specialist detectors report zero-shot performance on the full LVIS dataset, LLMs typically avoid this due to its large number of classes. We find that specialist detectors like GroundingDINO \cite{liu2023grounding}, SAM3 \cite{carion2025sam}, and LLMDet \cite{fu2025llmdet} consistently outperform our Qwen3-VL (30B-A3B) generalist baseline. Interestingly, zero-shot predictions from SAM3 slightly outperform GroundingDINO fine-tuned on LVIS v1, suggesting that LVIS categories are likely in-distribution for SAM3's PCS training dataset. Furthermore, LLMDet (which shares GroundingDINO's architecture but does not fine-tune on LVIS) achieves significantly lower performance. Consistent with the results in our main paper, DetPO yields a significant 3.2 mAP improvement over the Qwen3-VL baseline. VQA score further improves performance by 3.4 mAP. In contrast, prompt optimization techniques like GEPA and MIPROv2 fail to improve upon the Qwen3-VL's baseline prompt. We attribute this failure to such methods attempting to optimize for all 50 classes in a single prompt.

{
\setlength{\tabcolsep}{3em}
\begin{table}[H]
\centering
\caption{\textbf{LVIS Rare 50 10-Shot Benchmark}. We evaluate model performance on the 50 least frequent LVIS validation classes with atleast 10 examples per-class. While specialist models achieve the highest overall mAP, our DetPO approach and VQA Score substantially improve the performance of the generalist Qwen3-VL baseline.
} 
\label{tab:lvis}
\scalebox{0.70}{
\begin{tabular}{|lc|}
\hline

\rowcolor{gray!10}
\multicolumn{1}{|l|}{\textbf{Method}}                     &  \textbf{mAP} \\ \hline
\rowcolor{gray!10}
\multicolumn{2}{|l|}{\textbf{Specialist Models}}  \\  \hline 
\multicolumn{1}{|l|}{GroundingDINO (Fine-Tuned) \cite{liu2024grounding} (C)}  &       40.3       \\ \hline
\multicolumn{1}{|l|}{SAM3 \cite{carion2025sam} (C)}           &       \textbf{40.5}       \\ \hline
\multicolumn{1}{|l|}{LLMDet \cite{fu2025llmdet} (C)}         &         27.1     \\ \hline
\rowcolor{gray!10}
\multicolumn{2}{|l|}{\textbf{Generalist Models}}  \\  \hline 
\multicolumn{1}{|l|}{Qwen3-VL \cite{Qwen3-VL} (30B-A3B) (C + I)}      &      21.9        \\ \hline
\multicolumn{1}{|l|}{\quad w/ GEPA \cite{agrawal2025gepa}}     &     21.9        \\  \hline
\multicolumn{1}{|l|}{\quad w/ MIPROv2 \cite{opsahl2024optimizing}}     &    21.9         \\ \hline
\multicolumn{1}{|l|}{\quad w/ DetPO (Ours)}      &          25.1    \\ \hline
\multicolumn{1}{|l|}{\quad \quad + VQA Score}     &       \textbf{28.5}      \\ \hline
% \multicolumn{1}{|l|}{\quad w/ GEPA \cite{agrawal2025gepa}}     &     21.9        \\  \hline
% \multicolumn{1}{|l|}{\quad w/ MIPROv2 \cite{opsahl2024optimizing}}     &    21.9         \\ \hline
\end{tabular}
}
\end{table}
}

\begin{figure}[h]
    \centering
    \includegraphics[width=\linewidth]{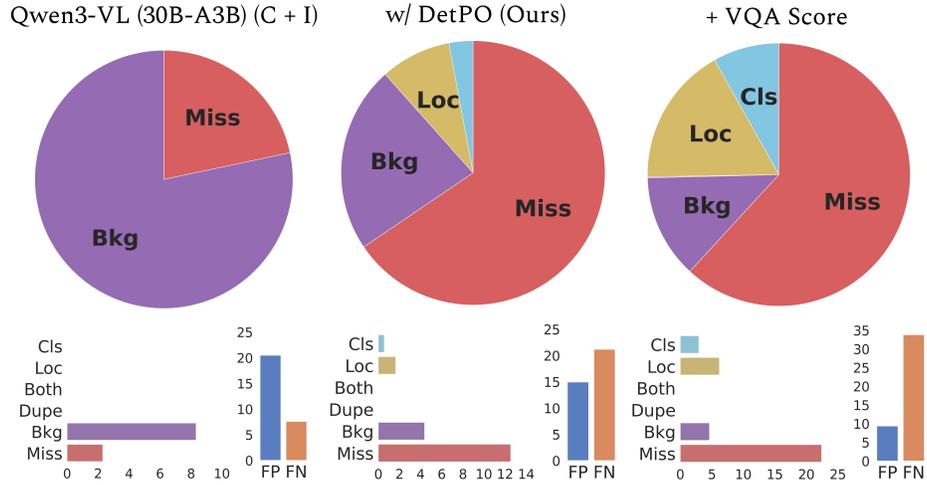}
    \caption{\textbf{Detection Errors.} We diagnose errors in the baseline Qwen3-VL (30B-A3B) model (\textbf{left}), the proposed DetPO method (\textbf{center}), and DetPO + VQA Score (\textbf{right}) with TIDE \cite{bolya2020tide}. The top row shows the relative distribution of error types, while the bottom row describes the absolute error counts and the overall false positive (FP) versus false negative (FN) rates. DetPO successfully reduces background errors (FP) at the cost of increased misses (FN), a trade-off that is further amplified when incorporating VQA Score.}
    \label{fig:lvis_detpo_erros}
\end{figure}

We further diagnose Qwen3-VL's errors in Figure \ref{fig:lvis_detpo_erros}. Comparing the baseline Qwen3-VL (left) to our DetPO method (center) demonstrates a significant reduction in background errors and overall false positives. While this trade-off inherently increases false negatives, our method yields a more balanced distribution of error types. The addition of VQA Score (right) further limits false positives but considerably increases false negatives.

\newpage

\section{Token and Run-Time Analysis}
\begin{figure}[t]
    \centering
    \includegraphics[width=0.85\linewidth]{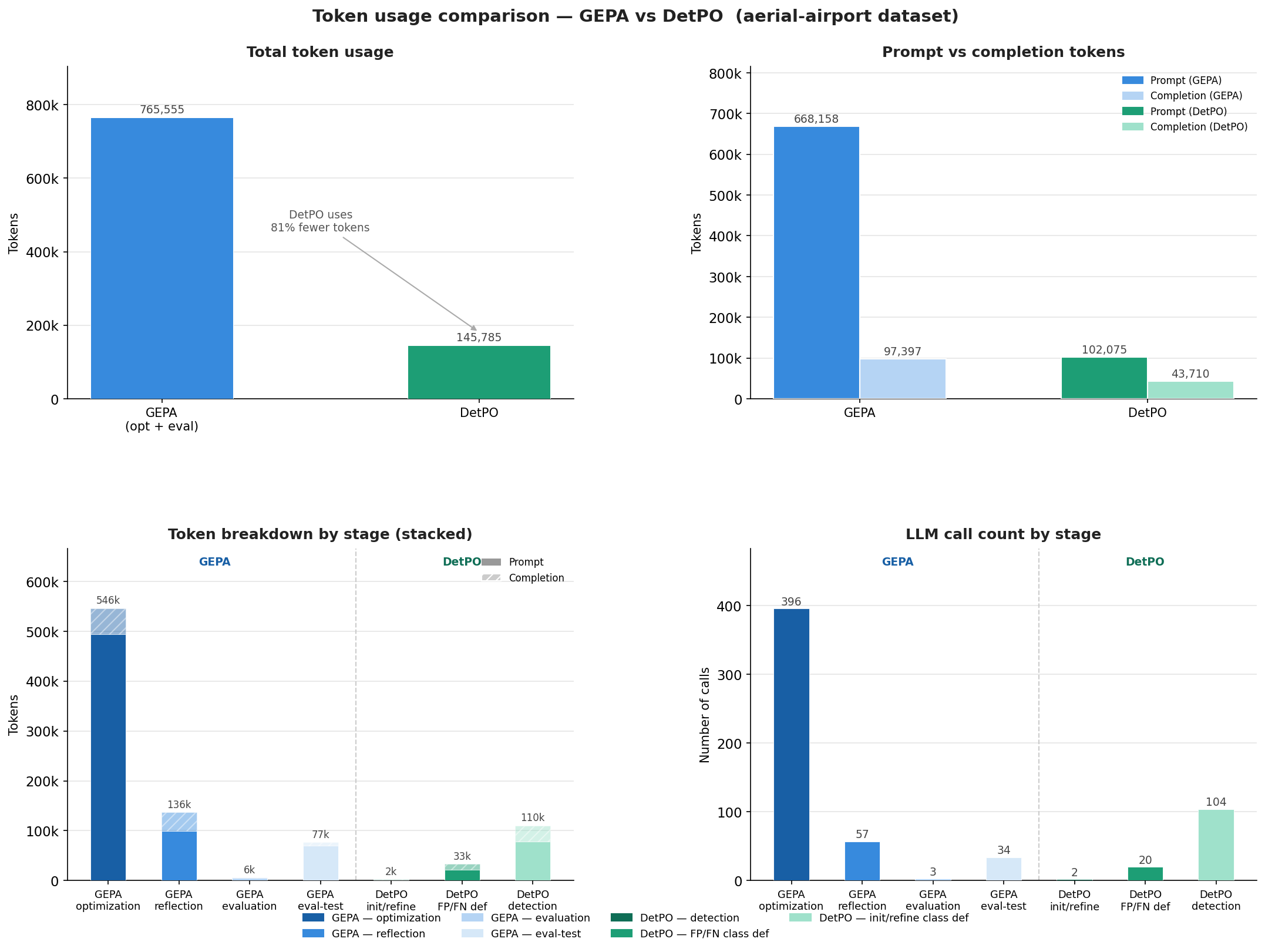}
    \caption{\textbf{Token Analysis.} We compare token usage between GEPA and DetPO on the aerial-airport dataset. DetPO uses 81\% fewer total tokens compared to GEPA (\textbf{top left}). DetPO's efficiency gains primarily stem from a massive reduction in prompt tokens (\textbf{top right}). In contrast, GEPA expends the vast majority of its tokens (546k) during the optimization phase, whereas DetPO's token usage is concentrated mostly in the final detection stage (\textbf{bottom left}). DetPO significantly reduces the number of required API calls compared to GEPA's intensive optimization loop (\textbf{bottom right}).}
    \label{fig:token_analysis}
\end{figure}

\begin{figure*}[!ht]
    \centering
    \begin{minipage}{0.65\textwidth}
        \centering
        \includegraphics[width=\linewidth]{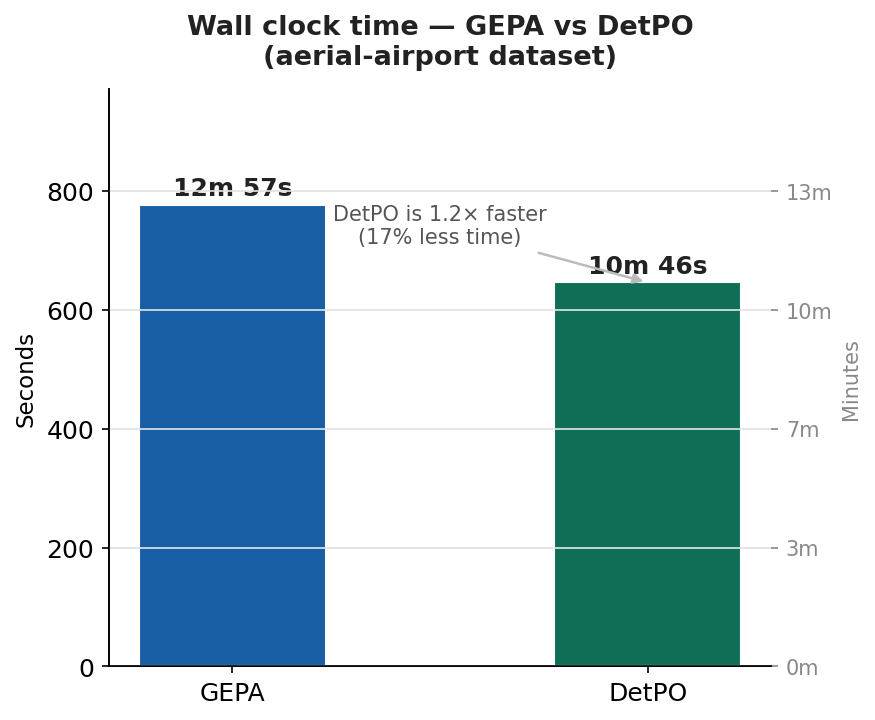}
    \end{minipage}
    \hfill
    \begin{minipage}{0.32\textwidth}
        \caption{\textbf{Wall Clock Analysis.} We compare the total execution time between GEPA and DetPO on the aerial-airport dataset. DetPO completes the task in 10 minutes and 46 seconds, outperforming GEPA's time of 12 minutes and 57 seconds. Overall, DetPO achieves a 17\% reduction in processing time, making it 1.2x faster than GEPA.}
        \label{fig:wallclock_analysis}
    \end{minipage}
\end{figure*}

Figures \ref{fig:token_analysis} and \ref{fig:wallclock_analysis} demonstrate the significant efficiency gains of DetPO over GEPA when evaluated on the aerial-airport dataset. As shown in Figure \ref{fig:token_analysis}, DetPO achieves an 81\% reduction in total token usage. Unlike GEPA, which exhausts the majority of its tokens during the optimization phase, DetPO bypasses this loop and concentrates its token usage almost entirely in the final detection stage. Consequently, this  reduction in LLM API calls and token consumption directly translates to faster execution. Figure \ref{fig:wallclock_analysis} illustrates that DetPO completes the task in 10 minutes and 46 seconds, making it 1.2$\times$ faster than GEPA's 12 minutes and 57 seconds runtime. Ultimately, DetPO offers a streamlined pipeline that conserves both computational resources and real-world processing time. Note that since GEPA optimizes multiple classes in parallel (unlike DetPO, which  optimizes each class separately), we expect that DetPO's runtime will increase proportional to the number of classes. Future work should parallelize DetPO's optimization and inference steps to further increase efficiency.

\newpage
\section{Impact of $K$ Shots}
\label{apdx:few_shot}
We evaluate the impact of the number of examples provided to DetPO during optimization in Table~\ref{tab:ablation_fewshot}. We observe that performance consistently improves when increasing from 3-shot to 5-shot settings. However, gains from 5-shot to 10-shot are marginal, suggesting diminishing returns and saturation of performance with additional training examples.

\vspace{-1em}

{
\setlength{\tabcolsep}{1.1em}
\begin{table}[h]
\centering
\caption{\textbf{Ablation on Few-Shot Examples}. We ablate the impact of the number of few-shot examples (3-shot, 5-shot, and 10-shot) using the DetPO on optimization performance. Notably, we see consistent improvements in increasing from 3 shots to 5 shots, but note marginal improvements from 5 shots to 10 shots, suggesting that adding additional examples does not significantly improve model performance. 
} 
\label{tab:ablation_fewshot}
\scalebox{0.70}{
\begin{tabular}{|lcccccccc|}
\hline

\rowcolor{gray!10}
\multicolumn{1}{|l|}{\textbf{Method}}                     & \multicolumn{1}{c|}{\textbf{A}} & \multicolumn{1}{c|}{\textbf{D}} & \multicolumn{1}{c|}{\textbf{F \& F}} & \multicolumn{1}{c|}{\textbf{I}} & \multicolumn{1}{c|}{\textbf{M}} & \multicolumn{1}{c|}{\textbf{S}} & \multicolumn{1}{c|}{\textbf{O}} & \textbf{All} \\ \hline
\multicolumn{1}{|l|}{Qwen3-VL \cite{Qwen3-VL} (30B-A3B) (C + I)}      & \multicolumn{1}{c|}{9.0}      & \multicolumn{1}{c|}{7.8}      & \multicolumn{1}{c|}{23.5}         & \multicolumn{1}{c|}{9.6}                    & \multicolumn{1}{c|}{0.7}                 & \multicolumn{1}{c|}{14.4}                & \multicolumn{1}{c|}{10.1}               &      11.9        \\ \hline
\hline
\multicolumn{1}{|l|}{\quad w/ DetPO 3-shot}      & \multicolumn{1}{c|}{12.9}      & \multicolumn{1}{c|}{17.2}      & \multicolumn{1}{c|}{33.4}         & \multicolumn{1}{c|}{19.1}                    & \multicolumn{1}{c|}{0.1}                 & \multicolumn{1}{c|}{22.1}                & \multicolumn{1}{c|}{16.4}               &      18.9        \\ \hline
% \multicolumn{1}{|l|}{\quad \quad + VQA Score}       & \multicolumn{1}{c|}{16.3}      & \multicolumn{1}{c|}{}      & \multicolumn{1}{c|}{46.1}         & \multicolumn{1}{c|}{}                    & \multicolumn{1}{c|}{0.1}                 & \multicolumn{1}{c|}{}                & \multicolumn{1}{c|}{}               &              \\ \hline
\multicolumn{1}{|l|}{\quad \quad + VQA Score}       & \multicolumn{1}{c|}{16.2}      & \multicolumn{1}{c|}{23.6}      & \multicolumn{1}{c|}{33.8}         & \multicolumn{1}{c|}{20.1}                    & \multicolumn{1}{c|}{0.1}                 & \multicolumn{1}{c|}{26.5}                & \multicolumn{1}{c|}{18.3}               &        21.0      \\ 
\hline
\hline
\multicolumn{1}{|l|}{\quad w/ DetPO 5-shot}      & \multicolumn{1}{c|}{14.1}      & \multicolumn{1}{c|}{18.1}      & \multicolumn{1}{c|}{34.5}         & \multicolumn{1}{c|}{19.0}                    & \multicolumn{1}{c|}{0.1}                 & \multicolumn{1}{c|}{22.9}                & \multicolumn{1}{c|}{15.6}               &      19.2        \\ \hline
\multicolumn{1}{|l|}{\quad \quad + VQA Score}       & \multicolumn{1}{c|}{16.4}      & \multicolumn{1}{c|}{25.6}      & \multicolumn{1}{c|}{36.6}         & \multicolumn{1}{c|}{19.7}                    & \multicolumn{1}{c|}{0.4}                 & \multicolumn{1}{c|}{26.1}                & \multicolumn{1}{c|}{17.9}               &     \textbf{21.6}         \\ \hline
\hline
\multicolumn{1}{|l|}{\quad w/ DetPO 10-shot}      & \multicolumn{1}{c|}{13.8}      & \multicolumn{1}{c|}{18.6}      & \multicolumn{1}{c|}{34.6}         & \multicolumn{1}{c|}{19.7}                    & \multicolumn{1}{c|}{0.1}                 & \multicolumn{1}{c|}{21.8}                & \multicolumn{1}{c|}{16.4}               &     19.4         \\ \hline
\multicolumn{1}{|l|}{\quad \quad + VQA Score \cite{lin2024evaluating}}      & \multicolumn{1}{c|}{16.1}      & \multicolumn{1}{c|}{25.2}      & \multicolumn{1}{c|}{36.5}         & \multicolumn{1}{c|}{20.1}                    & \multicolumn{1}{c|}{0.2}                 & \multicolumn{1}{c|}{25.7}                & \multicolumn{1}{c|}{18.4}               &     \textbf{21.6}         \\ \hline 
\end{tabular}
}
\end{table}
}

\vspace{-1em}

\section{Analysis on Prompt Optimization Variance} We evaluate DetPO's variance with Qwen3-VL (8B) over three runs in Table~\ref{tab:ipt_seed_variance}. Notably, DetPO's variance averaged over 20 datasets is 4.4\%, while its improvement over the baseline is 35.0\%. This shows that the improvements are consistent and significant. 

% % \vspace{-3mm}
% {
% \setlength{\tabcolsep}{0.5em}
% \begin{table}[h]
% \centering
% \scalebox{0.6}{
% \begin{tabular}{|l|c|c|c|c|}
% \hline
% \rowcolor{gray!10} 
% \textbf{Method} & \textbf{A} & \textbf{D} & \textbf{F \& F} & \textbf{I} \\ \hline
% & 8.40 $\pm$ 1.03 & 18.87 $\pm$ 0.69 & 30.55 $\pm$ 0.49 & 13.75 $\pm$ 0.21 \\ \cline{2-5} 
% \rowcolor{gray!10}
% \cellcolor{white} % Keeps the method name area white
% & \textbf{M} & \textbf{S} & \textbf{O} & \textbf{All} \\ \cline{2-5}
% \multirow{-3}{*}{Qwen3-VL (8B) + DetPO} & 0.10 $\pm$ 0.01 & 14.42 $\pm$ 0.50 & 11.98 $\pm$ 0.22 & 15.34 $\pm$ 0.36 \\ \hline
% \end{tabular}
% }
% \end{table}
% }

\vspace{-1em}

{
\setlength{\tabcolsep}{1.1em}
\begin{table}[h]
\centering
\caption{\textbf{Variance Analysis.} We evaluate DetPO's stability across three different runs with Qwen3-VL (8B). Scores remain consistent, with a relative variance of 4.4\%   averaged over 20 datasets.}
\label{tab:ipt_seed_variance}
\scalebox{0.6}{
\begin{tabular}{|lcccccccc|}
\hline
\rowcolor{gray!10}
\multicolumn{1}{|l|}{\textbf{Method}} & \multicolumn{1}{c|}{\textbf{A}} & \multicolumn{1}{c|}{\textbf{D}} & \multicolumn{1}{c|}{\textbf{F \& F}} & \multicolumn{1}{c|}{\textbf{I}} & \multicolumn{1}{c|}{\textbf{M}} & \multicolumn{1}{c|}{\textbf{S}} & \multicolumn{1}{c|}{\textbf{O}} & \textbf{All} \\ \hline
\rowcolor{gray!10}
\multicolumn{9}{|l|}{\textbf{Qwen3-VL-8B}} \\ \hline 
\multicolumn{1}{|l|}{\quad w/ DetPO (Seed 42)} & \multicolumn{1}{c|}{8.3} & \multicolumn{1}{c|}{19.1} & \multicolumn{1}{c|}{30.3} & \multicolumn{1}{c|}{13.7} & \multicolumn{1}{c|}{0.1} & \multicolumn{1}{c|}{14.2} & \multicolumn{1}{c|}{12.2} & 15.3 \\ \hline
\multicolumn{1}{|l|}{\quad w/ DetPO (Seed 43)} & \multicolumn{1}{c|}{9.5} & \multicolumn{1}{c|}{19.2} & \multicolumn{1}{c|}{30.7} & \multicolumn{1}{c|}{13.9} & \multicolumn{1}{c|}{0.1} & \multicolumn{1}{c|}{14.8} & \multicolumn{1}{c|}{12.0} & \textbf{15.6} \\ \hline
\multicolumn{1}{|l|}{\quad w/ DetPO (Seed 44)} & \multicolumn{1}{c|}{7.4} & \multicolumn{1}{c|}{18.9} & \multicolumn{1}{c|}{30.3} & \multicolumn{1}{c|}{14.1} & \multicolumn{1}{c|}{0.1} & \multicolumn{1}{c|}{15.0} & \multicolumn{1}{c|}{11.8} & 15.3 \\ \hline
\multicolumn{1}{|l|}{\textbf{Mean} $\pm$ \textbf{STD}} & \multicolumn{1}{c|}{8.4{$\pm$0.8}} & \multicolumn{1}{c|}{19.0{$\pm$0.1}} & \multicolumn{1}{c|}{30.4{$\pm$0.2}} & \multicolumn{1}{c|}{13.9{$\pm$0.2}} & \multicolumn{1}{c|}{0.1{$\pm$0.0}} & \multicolumn{1}{c|}{14.7{$\pm$0.4}} & \multicolumn{1}{c|}{12.0{$\pm$0.2}} & 15.4{$\pm$0.1} \\ \hline
\end{tabular}
}
\end{table}

}
\vspace{-1em}

% \newpage

\section{Ablation on Sampling Strategy}
We study the effect of the error selection strategy used during prompt refinement. Specifically, we compare our ``worst-case'' strategy, which selects the most severe false positives (i.e., highest-confidence false positives) and false negatives (i.e., lowest-IoU misses), against randomly sampled false positives and false negatives at each iteration. We find that prompts optimized using random samples perform marginally better at 10-shots (Table~\ref{tab:ipt_fpfn_selection}). However, this gap is within the inter-run variance. We hypothesize that, on larger training sets, prioritizing the most severe and representative failure cases encourages the model to capture corner cases and hard negatives, analogous to providing detailed corrective feedback to human annotators, whereas random errors would provide a weaker learning signal.

\vspace{-1em}

\setlength{\tabcolsep}{1.1em}
\begin{table}[h]
\centering
\caption{\textbf{FP/FN Selection Criterion.} We ablate different sampling strategies in selecting false positive and false negatives for prompt refinement. Notably, we find that selecting random errors performs marginally better in the 10-shot case than our proposed strategy, but posit that prioritizing the most severe and representative failure cases will yield greater benefits on larger training sets.}
\label{tab:ipt_fpfn_selection}
\scalebox{0.74}{
\begin{tabular}{|lcccccccc|}
\hline
\rowcolor{gray!10}
\multicolumn{1}{|l|}{\textbf{Method}} & \multicolumn{1}{c|}{\textbf{A}} & \multicolumn{1}{c|}{\textbf{D}} & \multicolumn{1}{c|}{\textbf{F \& F}} & \multicolumn{1}{c|}{\textbf{I}} & \multicolumn{1}{c|}{\textbf{M}} & \multicolumn{1}{c|}{\textbf{S}} & \multicolumn{1}{c|}{\textbf{O}} & \textbf{All} \\ \hline
\rowcolor{gray!10}
\multicolumn{9}{|l|}{\textbf{Qwen3-VL-8B}} \\ \hline 
\multicolumn{1}{|l|}{\quad w/ DetPO,\ Worst FP/FN} & \multicolumn{1}{c|}{8.3} & \multicolumn{1}{c|}{19.1} & \multicolumn{1}{c|}{30.3} & \multicolumn{1}{c|}{13.7} & \multicolumn{1}{c|}{0.1} & \multicolumn{1}{c|}{14.2} & \multicolumn{1}{c|}{12.2} & 15.3 \\ \hline
\multicolumn{1}{|l|}{\quad w/ DetPO,\ Random FP/FN} & \multicolumn{1}{c|}{9.2} & \multicolumn{1}{c|}{20.4} & \multicolumn{1}{c|}{30.2} & \multicolumn{1}{c|}{13.4} & \multicolumn{1}{c|}{0.2} & \multicolumn{1}{c|}{15.2} & \multicolumn{1}{c|}{12.2} & \textbf{15.6} \\ \hline
\rowcolor{gray!10}
\multicolumn{9}{|l|}{\textbf{Qwen3-VL-30B}} \\ \hline 
\multicolumn{1}{|l|}{\quad w/ DetPO,\ Worst FP/FN} & \multicolumn{1}{c|}{13.8} & \multicolumn{1}{c|}{18.6} & \multicolumn{1}{c|}{34.6} & \multicolumn{1}{c|}{19.7} & \multicolumn{1}{c|}{0.1} & \multicolumn{1}{c|}{21.8} & \multicolumn{1}{c|}{16.4} & 19.4 \\ \hline
\multicolumn{1}{|l|}{\quad w/ DetPO,\ Random FP/FN} & \multicolumn{1}{c|}{14.9} & \multicolumn{1}{c|}{18.9} & \multicolumn{1}{c|}{35.1} & \multicolumn{1}{c|}{18.7} & \multicolumn{1}{c|}{0.1} & \multicolumn{1}{c|}{21.5} & \multicolumn{1}{c|}{16.9} & \textbf{19.6} \\ \hline
\end{tabular}
}
\end{table}
\vspace{-1em}

\section{Adapting DetPO for Multiclass Evaluation}
We evaluate DetPO's per-class instructions under a multiclass evaluation setting. Directly concatenating all single-class prompts into a multi-class prompt leads to degraded performance due to excessive prompt length. Instead, we find that summarizing the per-class instructions into a compact unified prompt yields substantially better performance. As shown in Table~\ref{tab:ipt_single_vs_multi}, although DetPO's multi-class performance is lower than its single-class performance, this strategy still outperforms GEPA~\cite{agrawal2025gepa}. We provide the summarization prompt below.

{
\setlength{\tabcolsep}{1.1em}
\begin{table}[h]
\centering
\caption{\textbf{Single-class vs.\ Multi-class Evaluation.} We compare DetPO's single-class performance with its multi-class performance and GEPA's multi-class performance. Althought DetPO's multi-class performance is lower than its single-class performance, we find that it still outperforms GEPA~\cite{agrawal2025gepa}.}
\label{tab:ipt_single_vs_multi}
\scalebox{0.74}{
\begin{tabular}{|lcccccccc|}
\hline
\rowcolor{gray!10}
\multicolumn{1}{|l|}{\textbf{Method}} & \multicolumn{1}{c|}{\textbf{A}} & \multicolumn{1}{c|}{\textbf{D}} & \multicolumn{1}{c|}{\textbf{F \& F}} & \multicolumn{1}{c|}{\textbf{I}} & \multicolumn{1}{c|}{\textbf{M}} & \multicolumn{1}{c|}{\textbf{S}} & \multicolumn{1}{c|}{\textbf{O}} & \textbf{All} \\ \hline
\rowcolor{gray!10}
\multicolumn{9}{|l|}{\textbf{Qwen3-VL-8B}} \\ \hline 
\multicolumn{1}{|l|}{\quad w/ DetPO,\ Single-class} & \multicolumn{1}{c|}{8.3} & \multicolumn{1}{c|}{19.1} & \multicolumn{1}{c|}{30.3} & \multicolumn{1}{c|}{13.7} & \multicolumn{1}{c|}{0.1} & \multicolumn{1}{c|}{14.2} & \multicolumn{1}{c|}{12.2} & \textbf{15.3} \\ \hline
\multicolumn{1}{|l|}{\quad w/ DetPO,\ Multi-class} & \multicolumn{1}{c|}{8.6} & \multicolumn{1}{c|}{10.1} & \multicolumn{1}{c|}{26.9} & \multicolumn{1}{c|}{9.9} & \multicolumn{1}{c|}{0.1} & \multicolumn{1}{c|}{11.0} & \multicolumn{1}{c|}{10.5} & 12.5 \\ \hline
\multicolumn{1}{|l|}{\quad w/ GEPA \cite{agrawal2025gepa} (Multi-class)} & \multicolumn{1}{c|}{6.3} & \multicolumn{1}{c|}{10.8} & \multicolumn{1}{c|}{22.4} & \multicolumn{1}{c|}{8.5} & \multicolumn{1}{c|}{0.1} & \multicolumn{1}{c|}{12.8} & \multicolumn{1}{c|}{11.5} & 11.6 \\ \hline \hline
\rowcolor{gray!10}
\multicolumn{9}{|l|}{\textbf{Qwen3-VL-30B}} \\ \hline 
\multicolumn{1}{|l|}{\quad w/ DetPO, Single-class} & \multicolumn{1}{c|}{13.8} & \multicolumn{1}{c|}{18.6} & \multicolumn{1}{c|}{34.6} & \multicolumn{1}{c|}{19.7} & \multicolumn{1}{c|}{0.1} & \multicolumn{1}{c|}{21.8} & \multicolumn{1}{c|}{16.4} & \textbf{19.4} \\ \hline
\multicolumn{1}{|l|}{\quad w/ DetPO, Multi-class} & \multicolumn{1}{c|}{13.6} & \multicolumn{1}{c|}{12.0} & \multicolumn{1}{c|}{29.6} & \multicolumn{1}{c|}{11.7} & \multicolumn{1}{c|}{1.0} & \multicolumn{1}{c|}{17.5} & \multicolumn{1}{c|}{15.5} & 16.0 \\ \hline
\multicolumn{1}{|l|}{\quad w/ GEPA \cite{agrawal2025gepa} (Multi-class)}      & \multicolumn{1}{c|}{9.3}      & \multicolumn{1}{c|}{12.4}      & \multicolumn{1}{c|}{23.6}         & \multicolumn{1}{c|}{10.8}                    & \multicolumn{1}{c|}{1.3}                 & \multicolumn{1}{c|}{15.1}                & \multicolumn{1}{c|}{11.3}               &     13.0        \\ \hline
\end{tabular}
}
\end{table}
}
% prompt = (
%             f"You are building a reusable object-detection prompt for a dataset whose "
%             f"object classes are: {class_list_str}.\n"
%             f"Below are free-text descriptions, one per training image, of the objects a "
%             f"detector found in each image.\n"
%             f"Write a concise set of annotator-style instructions describing what these "
%             f"object classes look like and the visual cues useful for detecting them. "
%             f"Be specific and visual. Do not mention image numbers or that these came "
%             f"from descriptions.\n\n"
%             f"Descriptions:\n{joined}"
%         )

\newpage
\textbf{DetPO Multiclass Summarization Prompt}
\greybox{
\ttfamily\small
\texttt{You are building a reusable object-detection prompt for a dataset whose object classes are: \{class\_list\}.} \\
\texttt{Below are free-text descriptions of the objects a detector found from images.} \\
\texttt{Write a concise set of annotator-style instructions describing what these object classes look like and the visual cues useful for detecting them.} \\
\texttt{Be specific and visual. Do not mention image numbers or that these came from descriptions.} \\
\\
\texttt{Descriptions:} \\
\texttt{\{dataset\_instructions\}}
}

\end{document}